\documentclass[11pt]{article}
\usepackage{afin_preprint}

\usepackage{amsmath,amsfonts,bm}

\def\eqref#1{equation~\ref{#1}}

\def\1{\bm{1}}

\DeclareMathAlphabet{\mathsfit}{\encodingdefault}{\sfdefault}{m}{sl}
\SetMathAlphabet{\mathsfit}{bold}{\encodingdefault}{\sfdefault}{bx}{n}

\usepackage{hyperref}
\usepackage{url}

\usepackage[utf8]{inputenc} 
\usepackage{url}            
\usepackage{booktabs}       
\usepackage{amsfonts}       
\usepackage{nicefrac}       
\usepackage{microtype}      
\usepackage{graphicx}
\usepackage{amsmath}
\usepackage{algorithm}
\usepackage{algpseudocode}
\usepackage{enumitem}
\usepackage{soul} 
\usepackage{booktabs}
\usepackage{adjustbox}
\usepackage{array}
\usepackage[table]{xcolor}
\usepackage{pifont}
\usepackage{tabularx}
\usepackage{tabto}
\usepackage{makecell}
\usepackage{wrapfig}
\usepackage{float}
\usepackage[section]{placeins}
\usepackage{etoolbox}

\AtBeginEnvironment{table}{\small}
\newcommand{\cmark}{\textcolor{green!55!black}{\ding{51}}}

\newcommand{\pmark}{\textcolor{green!55!black}{\(\circ\)}} 

\newcommand{\lightrowrule}{%
  \arrayrulecolor{black!12}\specialrule{0.15pt}{0pt}{0pt}\arrayrulecolor{black}%
}

\usepackage{xcolor}

\definecolor{darkblue}{RGB}{0, 71, 153}  

\usepackage[nameinlink]{cleveref}

\title{Amortized Factor Inference Networks \\for Posterior Inference}

\author{Joohwan Ko \quad Justin Domke\\[0.3em]
{\normalfont\small Manning College of Information and Computer Sciences}\\
{\normalfont\small University of Massachusetts Amherst}\\
{\normalfont\small\texttt{\{joohwanko,domke\}@cs.umass.edu}}}

\begin{document}
\addtocontents{toc}{\protect\setcounter{tocdepth}{-1}}

\begin{tcolorbox}[afinfrontmatter]
\maketitle

\begin{abstract}
Amortized inference promises fast test-time Bayesian inference, but existing methods typically target a fixed model or a restricted model family.
Extending amortization to new model families often requires retraining or test-time fine-tuning.
In this paper, we ask: is it possible to build a single inference network capable of generalizing across varying priors, likelihoods, latent dimensions, and observation counts?
We introduce Amortized Factor Inference Networks (AFINs), which describe a model as a list of typed prior and likelihood factors and map this description, together with the observations, to a Gaussian posterior approximation through an encode--merge--decode network whose learned parameters do not depend on the latent dimension or the number of observations.
Experiments on synthetic and real-data benchmarks show that a single trained AFIN provides competitive posterior approximations at substantially lower test-time cost than NUTS and per-task variational inference.
Optional importance sampling can further improve accuracy.
\end{abstract}
\smallskip
{\raggedright\small\textbf{Code:} \url{https://github.com/joohwanko/AFINs}\par}
\end{tcolorbox}

\section{Introduction}


Simulation-based neural posterior estimation trains a network on simulated latent values and observations to predict an approximate posterior for a new dataset \citep{NIPS2016_6aca9700,pmlr-v97-greenberg19a,radev2022bayesflow,cranmer2020frontier}.
Maximum-likelihood training minimizes expected forward KL divergence from the true posterior to the approximation.
This favors coverage of posterior mass, which suits later refinement by importance sampling \citep{pmlr-v161-jerfel21a}.

However, training is expensive and usually specific to one model, so changing a prior or likelihood can require retraining.
Recent work extends amortization to varying observation counts \citep{reuter2025fullbayesicl,gloeckler2024allinone,vetter2025npepfn}, model parameters \citep{whittle2025distribution}, or probabilistic programs \citep{wu2022foundation}.
We ask whether one trained network can handle changes in prior and likelihood families, their parameters, latent dimensions, and observation counts jointly (see \Cref{tab:main-related} for a comparison with existing methods).

We introduce \emph{Amortized Factor Inference Networks} (AFINs), which take both the model and its observations as inputs.
We consider models with one global latent vector and observations that are conditionally independent given it.
Each model is described by one prior factor and one likelihood factor per observation, with a distribution type and numerical parameters for each factor.
Each network is trained on a chosen set of factor types.

Conjugate exponential-family updates suggest a simple way to organize inference: represent each factor by its natural-parameter contribution, add the contributions, and recover the posterior.
AFIN extends this idea beyond conjugate models with a learned \textbf{encoder} that maps each factor's inputs to an embedding, a \textbf{merge} that lets the embeddings interact through attention before summing them, and a \textbf{decoder} that maps the sum to a Gaussian posterior approximation.
The factor inputs and predicted posterior parameters also change size with the latent dimension.
We therefore use parameter-tied multilayer perceptrons (MLPs) and attention modules that share weights across latent coordinates \citep{zaheer2017deepsets,maron2019invariant}, allowing input and output dimensions to change without changing the learned parameter shapes.

To train AFIN across models, we sample a model, draw a latent value from its prior, and generate observations from its likelihoods.
Training maximizes the log density that the predicted posterior assigns to the latent value that generated those observations.
The latent value is known from simulation, so training requires no posterior estimates from another inference method.

Our contributions are:
\begin{itemize}[nosep,beginpenalty=10000]
    \item We introduce a factor-based approach that jointly amortizes inference across supported prior and likelihood families, parameter values, latent dimensions, and observation counts.
    \item We adapt permutation-equivariant MLPs and attention into an encode--merge--decode architecture whose learned parameter shapes are independent of the latent dimension and the number of observations.
    \item One trained AFIN produces posterior approximations in a single forward pass, matches much longer NUTS and variational runs after importance-sampling refinement at a fraction of their cost, extrapolates in the number of observations, and extends to a new likelihood type through a small adapter.
\end{itemize}

\begin{figure}[t]
    \centering
    \includegraphics[width=1.0\linewidth]{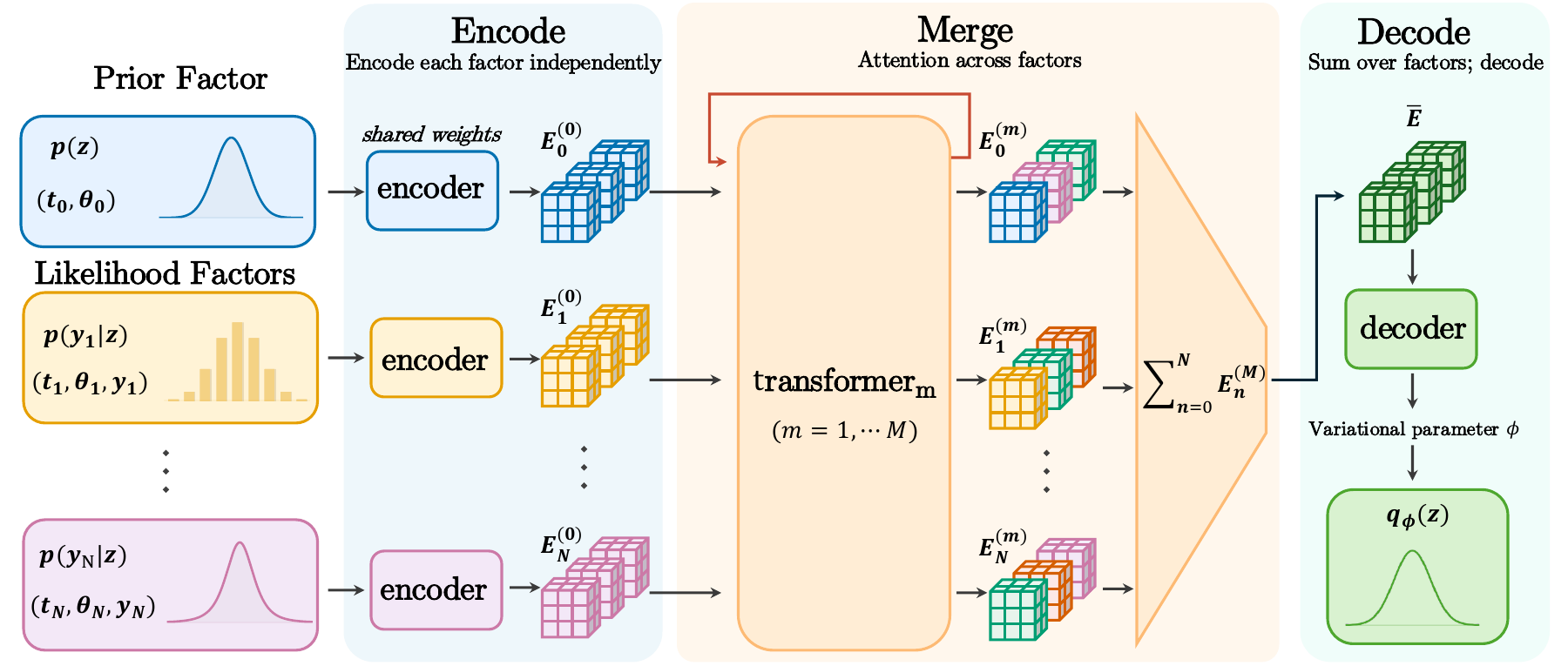}
    \caption{
        AFIN overview.
        The \textbf{encoder} maps the prior's type and parameters, and each likelihood's type, parameters, and observed value, to node and pair embeddings.
        The \textbf{merge} lets these embeddings exchange information through attention, then sums them over factors.
        The \textbf{decoder} maps the sum to the mean and precision of a Gaussian approximate posterior \(q_\phi(z)\).
        Sharing weights across latent coordinates allows the same trained network to handle different dimensions.
    }
    \label{fig:archi_diagram}
\end{figure}

\section{Problem setup}
\label{sec:problem-setting}

Standard neural posterior estimation amortizes inference over datasets from a fixed probabilistic model.
Let \(m\) denote the model, \(z\) its latent variables, and \(y\) its observations, with joint distribution \(p(z,y\mid m)\).
An inference network \(f_w\), with learned weights \(w\), maps \(y\) to the parameters \(\phi\) of an approximate posterior \(q(z;\phi)\).
The network is trained on simulated pairs \((z,y)\) with the objective
\begin{equation}
    \max_w\;\mathbb E_{p(z,y\mid m)}\left[\log q\bigl(z;f_w(y)\bigr)\right].
    \label{eq:fixed-model-objective}
\end{equation}
The aim is for \(q(z;f_w(y))\) to approximate \(p(z\mid y,m)\) for new observations from this same model.

We seek to amortize inference over models as well as datasets.
The network therefore receives both the observations \(y\) and the model \(m\).
During training, we sample \(m\) from a chosen distribution \(p(m)\), then simulate \((z,y)\) from that model, giving the objective
\begin{equation}
    \max_w\;\mathbb E_{p(m,z,y)}\left[\log q\bigl(z;f_w(y,m)\bigr)\right],
    \qquad p(m,z,y)=p(m)p(z,y\mid m).
    \label{eq:cross-model-objective}
\end{equation}
The aim is for the same learned weights \(w\) to yield accurate posterior approximations across the chosen range of models and datasets.
At inference time, \(m\) is known, and the target is \(p(z\mid y,m)\).

We focus on models with one shared latent vector \(z\in\mathbb R^D\) and \(N\) observations \(y=(y_1,\ldots,y_N)\) that are conditionally independent given \(z\).
We specify each model by \(m=(D,N,t_{0:N},\theta_{0:N})\), with joint density
\begin{equation}
    p(z,y\mid m)=p(z\mid t_0,\theta_0)\prod_{n=1}^N p(y_n\mid z,t_n,\theta_n),
    \label{eq:model-spec}
\end{equation}
where \((t_0,\theta_0)\) describes the prior factor and \((t_n,\theta_n)\) describes the likelihood factor for observation \(y_n\).
The label \(t_n\) identifies the factor type, such as a Gaussian prior or a Bernoulli-logit likelihood.
The known numerical parameters \(\theta_n\) specify the particular factor, including quantities such as means, scales, or covariates.
The type labels belong to a chosen finite set, and the parameters may include scalars, vectors, or matrices.
Different observations may have different likelihood types.

A model with \(D=1\) and \(N=3\) can combine continuous and binary observations as follows
\begin{alignat*}{3}
    z &\sim\mathcal N(0,\sigma_0^2),
    &\quad t_0&=\text{\small\texttt{"gaussian prior"}},
    &\quad\theta_0&=(0,\sigma_0),\\
    y_1 &\sim\mathcal N(z,\sigma_1^2),
    & t_1&=\text{\small\texttt{"gaussian likelihood"}},
    &\theta_1&=(\sigma_1),\\
    y_2 &\sim\mathrm{StudentT}_{\nu_2}(x_2z,\sigma_2),
    & t_2&=\text{\small\texttt{"student-t likelihood"}},
    &\theta_2&=(x_2,\nu_2,\sigma_2),\\
    y_3 &\sim\mathrm{Bernoulli}(\mathrm{sigmoid}(x_3z)),
    & t_3&=\text{\small\texttt{"bernoulli-logit likelihood"}},
    &\theta_3&=(x_3).
\end{alignat*}
Here, \(\sigma_0,\sigma_1,\sigma_2>0\) are scales, \(\nu_2>0\) is the degrees of freedom, and \(x_2,x_3\in\mathbb R\) are known covariates.
All three likelihoods depend on the same latent variable \(z\).

To define \(p(m)\), we choose how to sample the latent dimension, observation count, factor types, and their parameters; \Cref{sec:experimental-setup} describes these choices and the dimension-normalized form of \Cref{eq:cross-model-objective} that we optimize in practice.

The central challenge is to build one inference network whose learned parameters work across these variations: the number of factors changes with \(N\), their inputs differ by type, and the input and output dimensions change with \(D\).
The next section develops an architecture for this purpose.

\section{Amortized Factor Inference Networks}

\subsection{Architectural motivation}

Conjugate inference provides a simple template: summarize each factor, add the summaries, and recover the posterior from their sum.
To make this precise, consider an exponential family \(r(z\mid\eta)\propto\exp(\eta^\top T(z))\) with constant base measure.
Here, \(T(z)\) contains the sufficient statistics, and \(\eta\) contains their coefficients, called natural parameters.
Suppose the log prior and each log likelihood are linear in these statistics, up to terms independent of \(z\).
Denote their coefficient vectors by \(\eta_{t_0}(\theta_0)\) and \(\eta_{t_n}(\theta_n,y_n)\).
Multiplying the factors adds these coefficients, giving the posterior update
\begin{equation}
    p(z\mid y,m)=r\Bigl(z\Bigm|
    \eta_{t_0}(\theta_0)+\sum_{n=1}^N\eta_{t_n}(\theta_n,y_n)\Bigr).
    \label{eq:conjugate-update}
\end{equation}
Here, we \textbf{encode} each factor as its natural-parameter contribution, \textbf{merge} by addition, and \textbf{decode} the sum as a posterior distribution in the family \(r\).
For example, when the prior and all likelihoods are Gaussian in \(z\), each factor contributes a vector in \(\mathbb R^D\) and a matrix in \(\mathbb R^{D\times D}\), and the posterior is the Gaussian whose precision-weighted mean and precision are the sums of these contributions.

Beyond conjugate models, log factors still add, but two things break.
The sum need not correspond to any tractable family, and a tractable summary of one factor depends on where the other factors place their mass.
AFIN preserves the encode--merge--decode organization while learning all three steps: how to represent each factor, how to let factors exchange information before they are combined, and how to map the result to a posterior approximation.

\subsection{Architecture}
\label{sec:architecture}

The \textbf{encoder} maps each factor's inputs to an embedding, the \textbf{merge} lets the embeddings interact before summing them, and the \textbf{decoder} predicts posterior parameters \(\phi=f_w(y,m)\).
\Cref{fig:archi_diagram} shows these three steps.
This section explains what each step does and why; \Cref{sec:building-blocks} and \Cref{app:architecture-details} give the constructions.

\paragraph{Encoder.}
In a conjugate model, the natural parameters of a factor contain all the information about that factor needed for inference, in a form shared by every factor.
We want to preserve this uniform representation, since it is what allows a single merge step to combine factors of different types.
However, realistic models do not have simple low-dimensional sufficient statistics.
We therefore use high-dimensional embeddings of each factor, without requiring those embeddings to be the parameters of any simple family of distributions.
This gives the network the flexibility to represent whatever information is useful for combining factors, without building in a conjugate assumption.

What size should the embedding of a factor be?
For typical factors, the size of the parameters \(\theta_n\) grows linearly or quadratically with the latent dimension \(D\); a Gaussian prior, for example, has a mean vector and a precision matrix.
We therefore use an embedding \(E_n\) containing \emph{node features} for each latent coordinate and \emph{pair features} for each pair of coordinates
\begin{equation}
    E_n=\left(E_n^{\mathrm{node}},E_n^{\mathrm{pair}}\right),
    \qquad
    E_n^{\mathrm{node}}\in\mathbb R^{D\times C},
    \qquad
    E_n^{\mathrm{pair}}\in\mathbb R^{D\times D\times C},
    \label{eq:node-pair}
\end{equation}
where \(C\) is a fixed number of learned features per coordinate or pair.
For a Gaussian prior, the node features can hold the mean and the diagonal of the precision matrix, while the pair features can hold its off-diagonal entries, which describe dependence between coordinates.
The same shapes serve every factor type, so factors as different as a Gaussian prior and a Bernoulli-logit likelihood are represented in a common format.

A single embedding network cannot be applied to factors of varying type \(t_n\), because factor types differ in the number, shape, and meaning of the parameters \(\theta_n\) and observations \(y_n\).
One could learn a separate encoder for each factor type.
To reduce the number of type-specific parameters, we instead use a small type-specific \emph{adapter} that produces an initial embedding, followed by a larger network that is shared across all factor types.
The encoder output, the initial embedding of factor \(n\), is
\begin{equation}
    E_n^{(0)}
    =\mathtt{encoder}(t_n,\theta_n,y_n)
    =\mathtt{net}\bigl(\mathtt{adapter}_{t_n}(\theta_n,y_n)\bigr),
    \label{eq:encoder}
\end{equation}
where the prior factor has no observation and its adapter takes \(\theta_0\) alone.
The adapter handles the input format of its factor type, while the shared network performs most of the computation and lets information pass between node and pair features within the factor.
Both \(\mathtt{net}\) and \(\mathtt{adapter}_t\) share their learned parameters across latent coordinates, so they apply at any \(D\) (\Cref{sec:building-blocks}).

\paragraph{Merge.}
In a conjugate model, merging is addition: the natural parameters of all factors are summed, and nothing else is needed.
The direct analogue would be to sum the \(N+1\) factor embeddings.
We instead let the embeddings interact through attention before summing them, for two reasons.

First, factors depend on each other.
Outside conjugate models, a factor has no exact fixed-size summary, so its embedding is an approximation, and that approximation should be accurate where the posterior lies, which all factors decide together.
A sum of independently computed embeddings can capture this dependence only indirectly, through the decoder acting on the sum; attention lets each factor's embedding be updated with information from all others before the sum.

Second, self-attention without positional encodings is equivariant to the order of its tokens, and the final sum makes the merged embedding invariant to it, which matches the model: observations are conditionally independent given \(z\), so the posterior does not depend on their order.

We apply \(M\) Transformer blocks, \(\mathtt{transformer}_m\) in \Cref{fig:archi_diagram}, treating each factor's node and pair arrays as one token, and then sum the outputs to obtain \(\bar E=\sum_{n=0}^N E_n^{(M)}\).
Both node and pair features enter the attention weights, and the projections are parameter-tied MLPs preserving the node and pair shapes, so the same blocks operate at any \(D\) (\Cref{sec:boxtransformer}).
The sum removes the factor axis, giving the decoder one \(D\times C\) node array and one \(D\times D\times C\) pair array for any \(N\), and keeps the additive form of the conjugate update once the factors have exchanged information.

\paragraph{Decoder.}
In a conjugate model, the summed natural parameters are already the parameters of the posterior, which lies in the prior's family, so decoding is a fixed conversion.
Here, the merged embedding \(\bar E\) is a collection of learned features, not the parameters of a distribution.
The \(\mathtt{decoder}\) maps it to an approximate posterior \(q_\phi(z)=q(z;\phi)\) that supports the two operations we need: density evaluation for the training objective in \Cref{eq:cross-model-objective}, and sampling for posterior inference.
We use a Gaussian, \(q_\phi(z)=\mathcal N(z;\mu,\Lambda^{-1})\), with mean \(\mu\in\mathbb R^D\) and precision \(\Lambda\in\mathbb R^{D\times D}\).
Its parameters follow the structure of the embedding: shared MLPs predict the mean and the diagonal of \(\Lambda\) from per-coordinate features, and the off-diagonal entries of \(\Lambda\) from per-pair features.
To guarantee that \(\Lambda\) is positive definite, the decoder symmetrizes it and sets each diagonal entry above the sum of the absolute off-diagonal entries in its row.
This restricts \(q_\phi\) to Gaussians with strictly diagonally dominant precision matrices.
A Gaussian cannot represent every posterior, but it is fast to evaluate and sample, and \Cref{sec:inference} describes how importance sampling can correct it when needed.

Because shared weights preserve the association between features and coordinates throughout the network, the resulting Gaussian is equivariant to permutations of the latent coordinates: reordering the coordinates in the model description reorders the predicted mean and the rows and columns of the precision matrix in the same way (\Cref{app:architecture-details}).

\subsection{Inference}
\label{sec:inference}

Given a new model \(m=(D,N,t_{0:N},\theta_{0:N})\) and its observations \(y\), one forward pass of AFIN computes \(\phi=f_w(y,m)=(\mu,\Lambda)\), the mean and precision of \(q_\phi(z)=\mathcal N(z;\mu,\Lambda^{-1})\), which can be used directly as the final posterior approximation without any per-problem optimization.

Maximizing \Cref{eq:cross-model-objective} minimizes expected forward KL, which penalizes low density where the posterior has mass.
Standard ELBO-based VI instead minimizes reverse KL, so the two objectives can produce different approximations even within a Gaussian family \citep{blei2017variational}.
This mass-covering tendency motivates using \(q_\phi\) as an importance-sampling proposal \citep{pmlr-v161-jerfel21a}.

When further refinement is useful, we apply self-normalized importance sampling (SNIS) to samples from \(q_\phi\).
In our experiments, this can improve accuracy at modest additional test-time cost, requiring only sampling and density evaluations, with no further optimization.
Each sample is weighted by the prior times the likelihoods divided by the proposal density, and the normalized weights give posterior estimates.
We report direct predictions as \emph{AFIN} and the refinement as \emph{AFIN+SNIS}.

\section{Parameter Sharing Across Dimensions}
\label{sec:building-blocks}

The modules of \Cref{sec:architecture} must accept any latent dimension \(D\) with one fixed set of weights.
We achieve this by treating the latent coordinates like the elements of a set: the same computation is applied at every coordinate, and coordinates communicate only through averages, so no weight shape depends on \(D\), and the module is equivariant to permutations of the coordinates.

\paragraph{Parameter-tied MLPs.}
\label{sec:boxmlp}
For a map from \(x\in\mathbb R^D\) to \(o\in\mathbb R^D\), instead of \(D\)-dependent weight matrices, we compute, with \(x_i\) the \(i\)-th entry of \(x\) and \(j=1,\ldots,H\) indexing hidden units,
\begin{equation}
    h_{ij} = \rho\bigl(b_j+w_jx_i+\bar w_j\operatorname{mean}(x)\bigr),
    \qquad
    o_i = c+\sum_{j=1}^H v_jh_{ij},
\label{eq:boxmlp}
\end{equation}
where \(\rho\) is an activation function, \(\operatorname{mean}(x)\) averages the \(D\) entries, and the learned \(b,w,\bar w,v\in\mathbb R^H\) and \(c\in\mathbb R\) are shared by every coordinate.
The local term lets each output depend on its own input and the pooled term on all the others; this is the permutation-equivariant layer of Deep Sets \citep{zaheer2017deepsets}, with mean pooling.
AFIN applies the same principle to the \(D\times C\) node array and the \(D\times D\times C\) pair array of a factor: one MLP acts at every node and another at every pair, and each receives, besides its local features, a few \(C\)-dimensional averages of the rest of the factor, such as the average of its row of the pair array or the node features of its two coordinates.
These averages are a subset of the linear permutation-equivariant maps between vectors and matrices characterized by \citet{maron2019invariant}; \Cref{app:architecture-details} lists the exact inputs.

\paragraph{Transformer blocks.}
\label{sec:boxtransformer}
Attention exchanges information across factors, treating each factor's node and pair arrays as one token, in a standard pre-normalization Transformer block \citep{vaswani2017attention} whose query, key, value, output, and feed-forward maps are the parameter-tied MLPs above.
Attention scores average the query--key inner products over coordinates and coordinate pairs, and the resulting weights average the value tokens coordinate by coordinate, so factors mix while the coordinate structure inside each factor is untouched; without positional encodings the block is also equivariant to factor order.
Computation is asymptotically quadratic in \(D\) and in \(N\), and memory quadratic in \(D\) but linear in \(N\), while the learned parameters depend on neither (\Cref{app:architecture-details}).

\section{Related Work}
\label{sec:related-work}

\begin{table}[t]
\centering
\caption{Which aspects of a model can change at test time without retraining: observations \(y\), their number \(N\), latent dimension \(D\), factor parameters \(\theta\), and factor types \(t\) (\cmark{} supported, \pmark{} partial, - unsupported); the last column marks methods whose authors report NUTS-comparable accuracy (with importance sampling for AFIN) and is not a controlled comparison.}
\label{tab:main-related}
\footnotesize
\setlength{\tabcolsep}{4pt}
\begin{tabular}{lcccccc}
\toprule
Method & \(y\) & \(N\) & \(D\) & \(\theta\) & \(t\) & \(\approx\) NUTS \\
\midrule
Bayesian ICL \citep{reuter2025fullbayesicl}
& \cmark & \pmark & - & - & - & \cmark \\

Masked Language Inference \citep{wu2022foundation}
& \cmark & \cmark & \cmark & \cmark & \cmark & - \\

Masked Language Inference-Fit \citep{wu2022foundation}
& \cmark & - & - & - & - & \cmark \\

Simformer \citep{gloeckler2024allinone}
& \cmark & \pmark & - & - & - & \cmark \\

Distribution Transformers \citep{whittle2025distribution}
& \cmark & \cmark & - & \pmark & - & \cmark \\

NPE-PFN \citep{vetter2025npepfn}
& \cmark & \pmark & \pmark & - & - & \cmark \\
AFINs (ours)
& \cmark & \cmark & \cmark & \cmark & \cmark & \cmark \\
\bottomrule
\end{tabular}

\end{table}

Bayesian posterior inference is usually run per instance, by MCMC such as HMC and NUTS \citep{neal2011mcmc,JMLR:v15:hoffman14a,JSSv076i01} or by variational inference \citep{blei2017variational,hoffman2013stochastic,pmlr-v33-ranganath14,pmlr-v37-rezende15}.
Simulation-based inference learns from simulated pairs instead, from ABC to neural posterior, likelihood, and ratio estimators and globally amortized estimators such as BayesFlow and JANA \citep{beaumont2002approximate,cranmer2020frontier,NIPS2016_6aca9700,pmlr-v97-greenberg19a,pmlr-v89-papamakarios19a,pmlr-v119-hermans20a,pmlr-v119-durkan20a,Tejero-Cantero2020,pmlr-v130-lueckmann21a,radev2022bayesflow,radev2023jana}, but it amortizes over datasets from one simulator.
\Cref{tab:main-related} compares methods that widen the scope of one trained network by what can change at test time without retraining: a variable set of observations within a task family \citep{reuter2025fullbayesicl,gloeckler2024allinone,vetter2025npepfn}, additionally the prior \citep{whittle2025distribution}, or arbitrary probabilistic-program text for Masked Language Inference \citep{wu2022foundation}, whose NUTS-level accuracy however requires per-task fitting.
\Cref{app:related} discusses further related interfaces \citep{muller2022pfn,chang2025ace,elsemueller2024sensitivity,le2017inference,wu2020metaamortized}; AFIN differs in its input, an explicit list of typed factors with their parameters, which lets one network cover every column of \Cref{tab:main-related}, including NUTS-comparable accuracy, without per-task fitting. Of these, only NPE-PFN accepts our tasks without retraining, and \Cref{sec:amortized-baseline} compares with it.

Architecturally, the parameter-tied MLP of \Cref{sec:boxmlp} is the Deep Sets equivariant layer \citep{zaheer2017deepsets}, its node and pair summaries are a subset of the equivariant maps of \citet{maron2019invariant}, and attention over a set without positional encodings is the Set Transformer \citep{lee2019set}, extended to several sets by \citet{selby2022multiset}.

\section{Experiments}
\label{sec:experiments}
\setcounter{topnumber}{1}

All results use one trained AFIN; only the new-likelihood study trains further.
\Cref{app:simulator,app:evaluation-protocol,app:additional-results} give the setup and further results.

\begin{figure}[t]
    \centering
    \includegraphics[width=\linewidth]{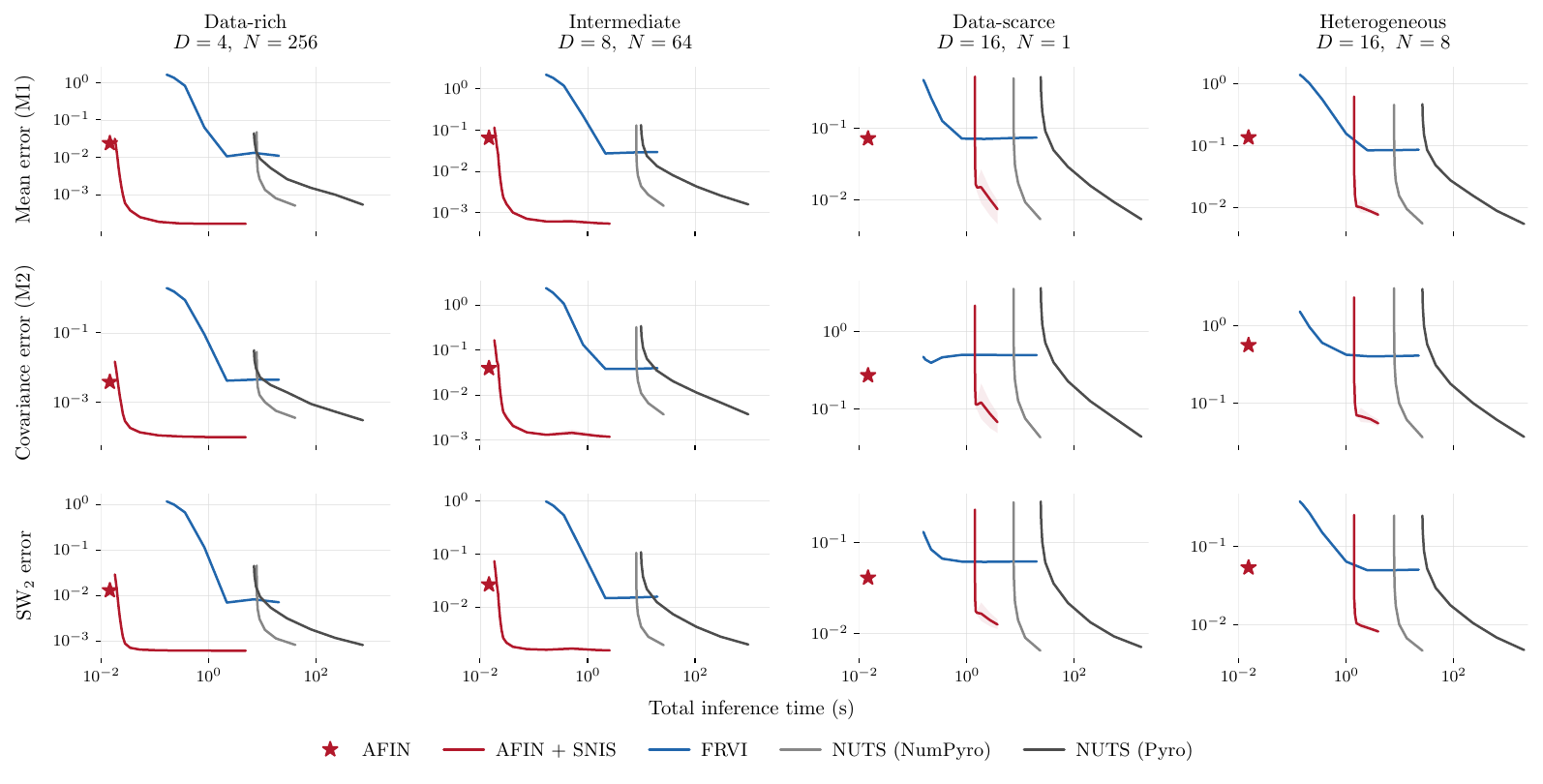}
    \caption{\textbf{AFIN gives a fast initial posterior, and importance sampling improves it without optimization.}
        Posterior error against total inference time, shown separately for three model sizes and heterogeneous likelihoods.
        Each of the first three columns averages 48 tasks; the last averages 36.
        Stars mark direct AFIN predictions; curves increase the sampling or optimization budget.
        SNIS curves use up to 500 million proposal samples per task.}
    \label{fig:synth-posterior-time}
\end{figure}

\subsection{Setup}
\label{sec:experimental-setup}

\begin{figure}[t]
    \centering
    \includegraphics[width=\linewidth]{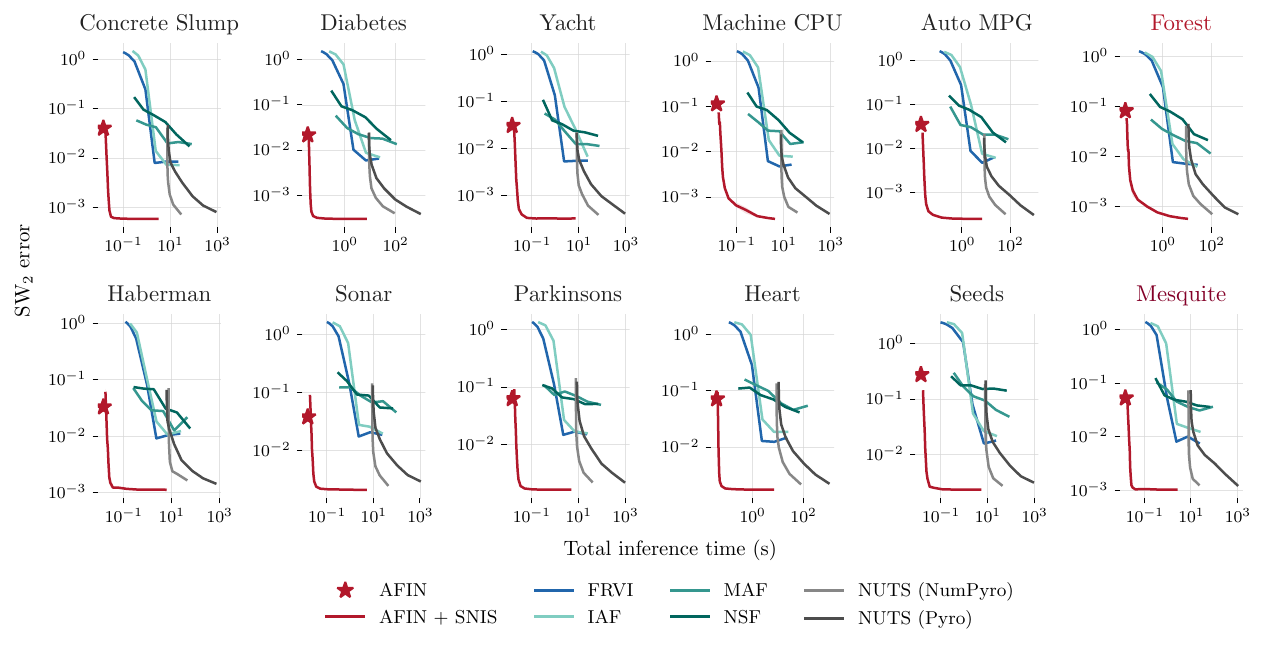}
    \caption{\textbf{On real data, AFIN with importance sampling reaches lower error than the baselines at the largest budgets shown on all 12 tasks.}
        Sliced Wasserstein-2 distance to a long-run NUTS reference against inference time on 12 UCI tasks.
        SNIS curves use up to 500 million proposal samples per task.
        Red titles mark models that mix two likelihood types; six tasks have \(N>256\), beyond the training range.}
    \label{fig:uci-results}
\end{figure}

\begin{figure}[t]
    \centering
    \includegraphics[width=\linewidth]{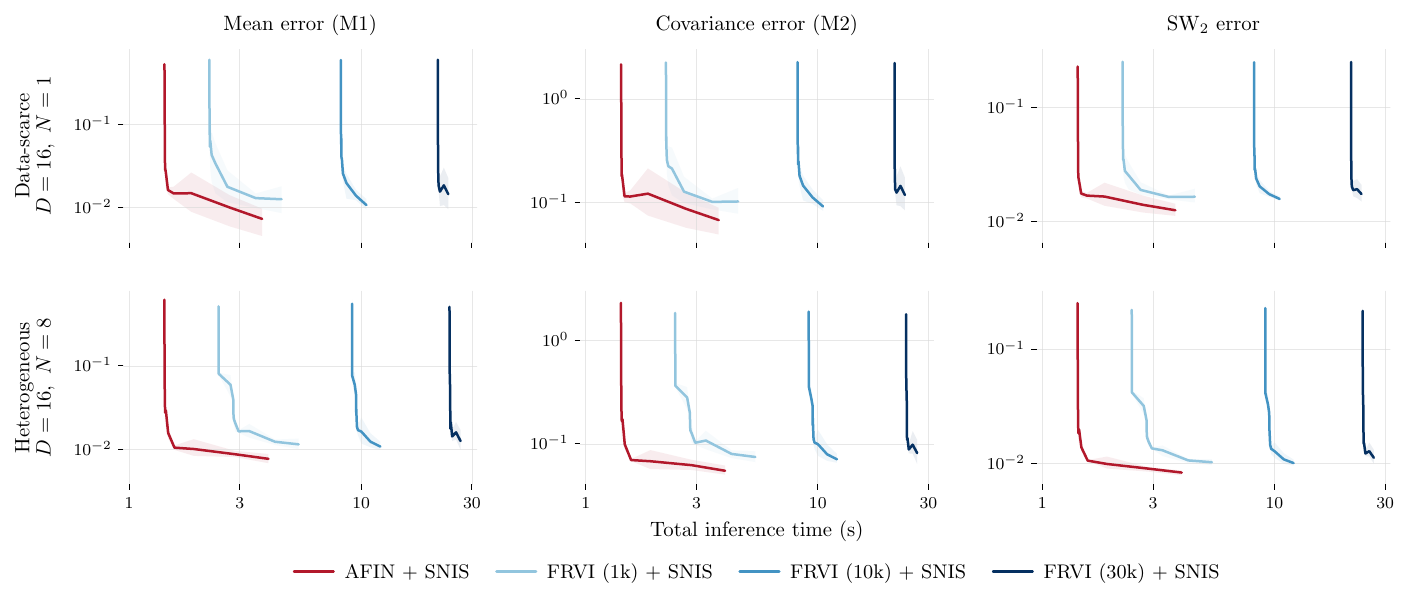}
    \caption{\textbf{AFIN is an effective importance-sampling proposal on data-scarce and heterogeneous tasks.}
        SNIS with AFIN or FRVI fitted for 1k, 10k, or 30k steps, using up to 500 million proposal samples per task.
        Time includes constructing or optimizing the proposal.
        Rows average 48 data-scarce tasks and 36 heterogeneous tasks; the remaining settings appear in \Cref{fig:snis-supporting,fig:heterogeneous-n32}.}
    \label{fig:synth-frvi-snis}
\end{figure}

\paragraph{Training.}
We train one \(4.55\)M-parameter AFIN with a Gaussian decoder on freshly simulated tasks with \(D\in\{1,\ldots,16\}\) and \(N\in\{1,\ldots,256\}\), sampled with extra weight on large \(D\) and small \(N\).
Prior types are \texttt{gaussian prior}, \texttt{student-t prior}, and \texttt{laplace prior}; likelihood types are \texttt{gaussian likelihood}, \texttt{student-t likelihood}, \texttt{bernoulli-logit likelihood}, and \texttt{binomial-logit likelihood}, and half of the models with more than one observation mix several of them.
We minimize the dimension-normalized loss
\begin{equation}
    \mathcal L(w)=\mathbb E_{p(m,z,y)}\bigl[-\tfrac{1}{D}\log q(z;f_w(y,m))\bigr].
    \label{eq:training_loss}
\end{equation}
The log density of a \(D\)-dimensional Gaussian is a sum over \(D\) coordinates, so without the \(1/D\) the highest-dimensional tasks would dominate the loss; dividing by \(D\) puts every task on a per-coordinate scale.
Since the simulated \(z\) is an exact posterior draw given \((m,y)\), this loss minimizes the expected forward KL divergence weighted by \(1/D\), with no reference posteriors needed.
We train for \(100{,}000\) AdamW steps \citep{loshchilov2019decoupled} (batch size \(32\), four-step gradient accumulation, cosine decay from \(2\times10^{-4}\)), under 24 hours on one H100 GPU, and evaluate exponential-moving-average weights.
\paragraph{Baselines, references, and metrics.}
We compare with the No-U-Turn Sampler (NUTS; \citealp{JMLR:v15:hoffman14a}), full-rank Gaussian variational inference (FRVI), and on real data flow-based variational inference (IAF, MAF, and NSF; \citealp{kingma2016iaf,papamakarios2017maf,durkan2019nsf}).
We include Pyro \citep{bingham2019pyro} to match the PyTorch backend of AFIN and VI, and NumPyro \citep{phan2019numpyro} for its faster JAX-based NUTS implementation.
For \Cref{fig:synth-posterior-time,fig:uci-results,fig:synth-frvi-snis}, we optimize each VI method at seven learning rates on separate validation tasks, then use the rate with the highest mean ELBO per latent dimension for all benchmark tasks.
AFIN and SNIS run on an A100 GPU; VI and NUTS run on one CPU core, following preliminary CPU/GPU comparisons that favored CPU execution.
Timings include setup and NUTS warmup (\Cref{app:baselines}).
We report posterior mean error (M1), covariance error (M2), and sliced Wasserstein-2 distance (SW\(_2\)) to reference posteriors; lower is better.
Each reference combines four long NumPyro NUTS chains; sampling budgets and mixing diagnostics are given in \Cref{app:references}.
SNIS shading shows variation across three sampling streams, not uncertainty across tasks.

\subsection{Accuracy and cost across model families}
\label{sec:synthetic-posterior}

\paragraph{Synthetic models.}
The first question is whether one network is accurate across the family it was trained on, and at what cost.
We cross four prior configurations with four likelihood types at three sizes: data-rich \((D=4,N=256)\), intermediate \((D=8,N=64)\), and data-scarce \((D=16,N=1)\), with three seeds per configuration (144 tasks).
We also test 36 heterogeneous tasks at \((D,N)=(16,8)\), mixing two or three supported likelihood types; \Cref{app:synthetic-tasks} gives their construction and an additional 36 tasks at \(N=32\).
In \Cref{fig:synth-posterior-time}, one AFIN forward pass takes about 15 ms and gives errors of the same order as the longer FRVI fits.
SNIS improves these predictions quickly; at larger time budgets, NUTS is more accurate on the data-scarce and heterogeneous tasks.
In a separate calibration study, AFIN's 90\% intervals cover \(0.875\) of the true coordinates in distribution (\Cref{app:calibration}).

\paragraph{Real data.}
\label{sec:uci}
The second question is whether it transfers to real datasets, including sizes beyond its training range.
The UCI benchmark has 12 Bayesian regression and classification tasks (\Cref{tab:uci-task-specs}); two mix a continuous and a binary likelihood in one model, and six have \(N>256\), beyond the training range.
In \Cref{fig:uci-results}, AFIN provides posterior estimates across all 12 tasks in a single forward pass, without per-task optimization.
Optional SNIS refinement further improves these estimates: with 500 million proposals, AFIN+SNIS has lower SW\(_2\) than every baseline at its largest shown budget, in about 2--10 seconds per task.

\subsection{Importance sampling refinement}
\label{sec:snis-refinement}

Forward-KL training penalizes low proposal density in regions with posterior mass, which can help importance sampling.
\Cref{fig:synth-frvi-snis} compares AFIN and reverse-KL FRVI under the same SNIS budgets, including the cost of building each proposal.
On data-scarce and heterogeneous tasks, AFIN+SNIS ends with lower errors and less total time; optimizing FRVI longer does not consistently improve its proposal.
On data-rich and intermediate tasks, the final errors are similar (\Cref{fig:snis-supporting}).

\subsection{Beyond the training distribution}
\label{sec:extrapolation}
\label{sec:factor-extension}

\paragraph{A new likelihood type.}
Each factor type enters through its own adapter, so a type the network has never seen can be added by training that adapter alone.
We test this with a Poisson likelihood with a log link, unseen in pretraining, training either a new adapter with the network frozen or adapter and network jointly, and evaluate on 24 held-out Poisson tasks (\Cref{tab:poisson-extension}).
Full fine-tuning is more accurate directly, while adapter-only training costs about 60\% of the time and leaves every original factor type untouched; after SNIS the two are indistinguishable.
A new likelihood therefore costs one adapter and about five GPU-hours.

\begin{table}[H]
\centering
\small
\caption{Adding a Poisson likelihood: errors over 24 tasks and three seeds, direct and after SNIS (4,096 proposals); best in bold.}
\label{tab:poisson-extension}
\begin{tabular}{@{}lrrrr@{}}
\toprule
Training mode & M1 & M2 & SW$_2$ & GPU-h \\
\midrule
Adapter only & 0.117 & 0.156 & 0.0591 & \textbf{4.95} \\
\multicolumn{1}{@{}r}{+SNIS} & \textbf{0.0319} & \textbf{0.113} & \textbf{0.0301} & \\
\addlinespace[2pt]
Full fine-tuning & 0.0785 & 0.140 & 0.0458 & 8.39 \\
\multicolumn{1}{@{}r}{+SNIS} & 0.0331 & 0.115 & 0.0305 & \\
\bottomrule
\end{tabular}
\end{table}

\paragraph{Larger models and datasets.}
Does it hold up beyond the sizes it trained on?
\Cref{tab:extrapolate-highdn} evaluates the same checkpoint on six configurations with \(D\le32\) and \(N\le512\), grouped by whether \(D\), \(N\), or both exceed the training ranges of \(16\) and \(256\).
When only \(N\) grows, or both grow, AFIN+SNIS has the lowest mean error on all three metrics, within noise of the best baseline on several, in under half a second against 32--157 s for FRVI and NUTS.
Growing only \(D\) is harder: the direct prediction degrades, SNIS lowers the mean and SW\(_2\) errors but raises the covariance error, and NUTS is most accurate at far larger cost.
Separate calibration tests give 90\% coverage of \(0.91\) at \(N\le400\) and \(0.57\) at \(N\le1{,}024\).

\begin{table}[H]
\centering
\small
\renewcommand{\arraystretch}{1.05}
\caption{Extrapolation in \(D\) and \(N\): SW\(_2\) (top), time per task (bottom); best in bold.}
\label{tab:extrapolate-highdn}
\begin{tabular}{@{}lcccc@{}}
\toprule
Split & AFIN & AFIN+SNIS & NUTS & FRVI \\
\midrule
OOD \(D\) & \makecell{0.031\\{\scriptsize 0.02 s}} & \makecell{0.024\\{\scriptsize 0.24 s}} & \makecell{\textbf{0.012}\\{\scriptsize 308 s}} & \makecell{0.042\\{\scriptsize 27 s}} \\
\addlinespace[2pt]
OOD \(N\) & \makecell{0.027\\{\scriptsize 0.02 s}} & \makecell{\textbf{0.0013}\\{\scriptsize 0.25 s}} & \makecell{0.0025\\{\scriptsize 95 s}} & \makecell{0.0025\\{\scriptsize 32 s}} \\
\addlinespace[2pt]
OOD \(D,N\) & \makecell{0.040\\{\scriptsize 0.10 s}} & \makecell{\textbf{0.0039}\\{\scriptsize 0.45 s}} & \makecell{0.0052\\{\scriptsize 157 s}} & \makecell{0.0041\\{\scriptsize 100 s}} \\
\bottomrule
\end{tabular}
\end{table}

\subsection{Amortized baseline}
\label{sec:amortized-baseline}

An in-context learner fed a task's simulations also needs no training; does it match AFIN?
NPE-PFN \citep{vetter2025npepfn} does this with a frozen tabular foundation model \citep{hollmann2025tabpfnv2}.
We ran it on the 144 synthetic tasks with 1k or 10k context simulations and 2,048 samples (\Cref{tab:npe-pfn}).
AFIN is more accurate on every metric in under a second; NPE-PFN takes 13 s, or 115 s with the larger context, which halves its mean error but not to AFIN's.

\begin{table}[H]
\centering
\small
\caption{NPE-PFN and AFIN on 144 synthetic tasks in a separate 2,048-sample comparison. Time includes inference and sampling; best in bold.}
\label{tab:npe-pfn}
\begin{tabular}{@{}lrrrr@{}}
\toprule
Method & M1 & M2 & SW$_2$ & s/task \\
\midrule
AFIN & \textbf{0.053} & \textbf{0.105} & \textbf{0.041} & \textbf{0.95} \\
NPE-PFN 1k & 0.242 & 0.301 & 0.113 & 13.2 \\
NPE-PFN 10k & 0.130 & 0.254 & 0.076 & 115 \\
\bottomrule
\end{tabular}
\end{table}

\subsection{Ablations}
\label{sec:ablations}

Do attention across factors and the pair array earn their place?
\Cref{tab:ablation} removes each: attention is replaced by a plain sum of the encoded factors, the conjugate template of \Cref{sec:architecture} with a learned encoder and decoder, and the pair array is dropped, leaving node features only.
Both matter, and their effects compound: dropping the pair array doubles the covariance error, since the decoder must then build the precision from node features alone, dropping attention triples SW\(_2\) and raises the mean error fivefold, and dropping both is worst on every metric.
The variants are smaller, so the gaps bound module and capacity together.
Further ablations and factor-composition experiments are in \Cref{app:additional-ablations}.

\begin{table}[H]
\centering
\small
\caption{Ablations on the synthetic suite: direct-prediction errors of variants trained with the same simulator and schedule; the first row is AFIN; best in bold.}
\label{tab:ablation}
\begin{tabular}{@{}llrrr@{}}
\toprule
Embedding & Merge & M1 & M2 & SW$_2$ \\
\midrule
Node+pair & Attention & \textbf{0.053} & \textbf{0.105} & \textbf{0.041} \\
Node only & Attention & 0.073 & 0.208 & 0.049 \\
Node+pair & Sum       & 0.290 & 0.248 & 0.132 \\
Node only & Sum       & 0.314 & 0.276 & 0.142 \\
\bottomrule
\end{tabular}
\end{table}

\section{Discussion}
\label{sec:discussion}

AFIN describes a model as a list of typed factors and turns one simulation-trained network into a posterior approximator for every model the list can express, with weights that depend on neither \(D\) nor \(N\).
One trained network gives a usable posterior in one forward pass across priors, likelihoods, and sizes, matches much longer NUTS and variational runs after importance sampling at a fraction of their cost, transfers to real data, gains a new likelihood through a cheap adapter, and beats an in-context foundation-model baseline.
Three limitations remain, each a direction for future work: the scope is one global latent vector with conditionally independent observations and known factor parameters, and
could extend to hierarchical models and local latent variables; the decoder is Gaussian, which limits the posteriors it can express, and a mixture or flow decoder would lift
this while keeping a density for SNIS; and cost grows quadratically with \(D\) and \(N\), with extrapolation in \(D\) degrading accuracy first,
which training at larger sizes would address.

\clearpage

\bibliographystyle{plainnat}
\bibliography{reference}

\clearpage
\appendix
\addtocontents{toc}{\protect\setcounter{tocdepth}{2}}
\begingroup
\renewcommand{\contentsname}{Appendix Contents}
\pdfbookmark[0]{Appendix Contents}{appendix-contents}
\tableofcontents
\endgroup
\clearpage
\setcounter{topnumber}{2}
\pretocmd{\subsection}{\FloatBarrier}{}{}
\section{Notation}
\label{app:notations}

\begin{table}[!htbp]
\centering
\caption{Main notation. The prior is factor \(n=0\) and the likelihoods are factors \(n=1,\ldots,N\).}
\label{tab:notation}
\small
\setlength{\tabcolsep}{5pt}
\renewcommand{\arraystretch}{1.12}
\begin{tabular}{@{}lp{0.6\linewidth}@{}}
\toprule
Symbol & Meaning \\
\midrule
\multicolumn{2}{@{}l}{\textit{Model}} \\
\midrule
\(z\in\mathbb R^D\) & Latent vector; \(D\) is the latent dimension. \\
\(y=(y_1,\ldots,y_N)\) & Observations, one likelihood factor each. \\
\(m=(D,N,t_{0:N},\theta_{0:N})\) & Model description: sizes, factor types, and factor parameters. \\
\(t_n\), \(\theta_n\) & Type label and known numerical parameters of factor \(n\). \\
\(p(z\mid t_0,\theta_0)\), \(p(y_n\mid z,t_n,\theta_n)\) & Prior and likelihood factors. \\
\midrule
\multicolumn{2}{@{}l}{\textit{Inference network}} \\
\midrule
\(f_w\), \(\phi=f_w(y,m)\) & AFIN with learned weights \(w\), and the posterior parameters it predicts. \\
\(q_\phi(z)=\mathcal N(z;\mu,\Lambda^{-1})\) & Gaussian posterior approximation with mean \(\mu\) and precision \(\Lambda\). \\
\(\mathcal L(w)\) & Training loss, \Cref{eq:training_loss}. \\
\(\mathtt{adapter}_t\), \(\mathtt{net}\) & Type-specific adapter and shared network that form the \(\mathtt{encoder}\). \\
\(E_n=(E_n^{\mathrm{node}},E_n^{\mathrm{pair}})\) & Embedding of factor \(n\): a \(D\times C\) node array and a \(D\times D\times C\) pair array. \\
\(C\) & Number of features per coordinate or pair. \\
\(\mathtt{transformer}_m\), \(M\) & \(m\)-th merge block and the number of blocks; \(E_n^{(m)}\) is the embedding after block \(m\). \\
\(\bar E=\sum_n E_n^{(M)}\) & Merged embedding passed to the \(\mathtt{decoder}\). \\
\midrule
\multicolumn{2}{@{}l}{\textit{Modules}} \\
\midrule
\(H\) & Hidden width of the coordinate-wise MLPs. \\
\(b_Q,b_K,b_V,b_{\mathrm{out}},b_{\mathrm{FFN}}\) & Node-pair maps inside a merge block. \\
\(Q_n,K_n,V_n\), \(\widetilde V_n\) & Query, key, value, and attended value of factor \(n\). \\
\(\mathsf h\), \(u\) & Number of attention heads and per-head width. \\
\(\operatorname{sym}\), \(\operatorname{LN}\) & Symmetrization of a pair array; layer normalization over features. \\
\midrule
\multicolumn{2}{@{}l}{\textit{Evaluation}} \\
\midrule
M1, M2, SW\(_2\) & Posterior mean error, covariance error, and sliced Wasserstein-2 distance, \Cref{app:metrics}. \\
SNIS & Self-normalized importance sampling with \(q_\phi\) as the proposal. \\
\bottomrule
\end{tabular}
\end{table}

\section{Architecture Details}
\label{app:architecture-details}

This appendix specifies the modules of \Cref{sec:architecture,sec:building-blocks}.
Every learned map is a small MLP applied with shared weights at each latent coordinate or coordinate pair, so no parameter shape depends on \(D\) or \(N\).

\paragraph{Node-pair embeddings.}
A factor embedding is a pair of arrays \(E=(E^{\mathrm{node}},E^{\mathrm{pair}})\) with \(E^{\mathrm{node}}\in\mathbb R^{D\times C}\) and \(E^{\mathrm{pair}}\in\mathbb R^{D\times D\times C}\).
We write \(E^{\mathrm{node}}_i\in\mathbb R^C\) for the features of coordinate \(i\) and \(E^{\mathrm{pair}}_{ij}\in\mathbb R^C\) for those of the pair \((i,j)\).
Pair arrays are kept symmetric by \(\operatorname{sym}(X)_{ij}=\tfrac12(X_{ij}+X_{ji})\) after every update.

\paragraph{Coordinate-wise MLPs.}
A coordinate-wise MLP \(g:\mathbb R^{C_{\mathrm{in}}}\to\mathbb R^{C_{\mathrm{out}}}\) is applied independently at every coordinate of a node array, \([g(X)]_i=g(X_i)\), or at every pair of a pair array, \([g(X)]_{ij}=g(X_{ij})\), with the same weights.
Unless stated otherwise, these MLPs have two hidden layers of width \(H=192\) with SiLU activations, and every embedding has \(C=40\) features.

\paragraph{Summaries.}
The parameter-tied maps combine local features with the following averages of a node-pair embedding, each in \(\mathbb R^C\),
\begin{equation}
\begin{aligned}
    \operatorname{row}(E^{\mathrm{pair}})_i &= \tfrac{1}{D}\textstyle\sum_{j} E^{\mathrm{pair}}_{ij}, &
    \operatorname{col}(E^{\mathrm{pair}})_j &= \tfrac{1}{D}\textstyle\sum_{i} E^{\mathrm{pair}}_{ij}, &
    \operatorname{diag}(E^{\mathrm{pair}})_i &= E^{\mathrm{pair}}_{ii},\\
    \overline{E}^{\mathrm{pair}} &= \tfrac{1}{D^{2}}\textstyle\sum_{i,j} E^{\mathrm{pair}}_{ij}, &
    \overline{E}^{\mathrm{node}} &= \tfrac{1}{D}\textstyle\sum_{i} E^{\mathrm{node}}_{i}.
\end{aligned}
\label{eq:summaries}
\end{equation}
For a symmetric pair array the row and column averages coincide; we keep both because the same maps are also applied to intermediate arrays that need not be symmetric.

\paragraph{Node-pair maps.}
A node-pair map takes a node-pair embedding and returns one.
It uses two coordinate-wise MLPs \(g^{\mathrm{node}},g^{\mathrm{pair}}:\mathbb R^{6C}\to\mathbb R^{C}\) whose inputs are
\begin{equation}
\begin{aligned}
    \zeta^{\mathrm{node}}_{i} &= \big[E^{\mathrm{node}}_{i},\;\operatorname{diag}(E^{\mathrm{pair}})_{i},\;\operatorname{row}(E^{\mathrm{pair}})_i,\;\operatorname{col}(E^{\mathrm{pair}})_i,\;\overline{E}^{\mathrm{pair}},\;\overline{E}^{\mathrm{node}}\big],\\
    \zeta^{\mathrm{pair}}_{ij} &= \big[E^{\mathrm{pair}}_{ij},\;\operatorname{row}(E^{\mathrm{pair}})_i,\;\operatorname{col}(E^{\mathrm{pair}})_j,\;E^{\mathrm{node}}_{i},\;E^{\mathrm{node}}_{j},\;\overline{E}^{\mathrm{pair}}\big],
\end{aligned}
\label{eq:np-inputs}
\end{equation}
and returns \(\bigl(g^{\mathrm{node}}(\zeta^{\mathrm{node}}),\,\operatorname{sym}(g^{\mathrm{pair}}(\zeta^{\mathrm{pair}}))\bigr)\).
A \emph{node-pair layer} adds this output to its input as a residual update and symmetrizes the pair array again.
Global averages are invariant to coordinate permutations, while row and column averages and local features move with their coordinates.
With shared MLP weights, both the map and the residual layer are therefore permutation-equivariant.

\subsection{Encoder}
\label{app:encoder}

\paragraph{Adapters.}
The adapter \(\mathtt{adapter}_{t}\) turns the numerical inputs of a factor of type \(t\) into an initial node-pair embedding.
It builds a descriptor vector for every coordinate, \(\xi^{\mathrm{node}}_{i}\in\mathbb R^{\chi^{\mathrm{node}}}\), and for every coordinate pair, \(\xi^{\mathrm{pair}}_{ij}\in\mathbb R^{\chi^{\mathrm{pair}}}\), whose lengths depend on the type but not on \(D\).
The descriptors pass through \(\operatorname{symlog}(x)=\operatorname{sign}(x)\log(1+|x|)\) and then through two coordinate-wise MLPs with one hidden layer of width \(64\) that output \(C\) features; the pair output is symmetrized.
Factor types with different input forms use different adapters.
A Gaussian prior with diagonal or full covariance, and a Gaussian likelihood with or without covariates, count as separate forms, giving four prior and five likelihood adapters in our experiments, with \(55{,}376\) parameters in total.
For example, a diagonal Gaussian prior \(z_i\sim\mathcal N(\mu_i,\sigma_i^2)\) uses
\begin{equation}
    \xi_i^{\mathrm{node}}=[\mu_i,\;\log\sigma_i,\;\mu_i/\sigma_i],
    \qquad
    \xi_{ij}^{\mathrm{pair}}=[\mu_i,\;\log\sigma_i,\;\mu_i/\sigma_i,\;\mu_j,\;\log\sigma_j,\;\mu_j/\sigma_j,\;\mathbf 1\{i=j\}],
\end{equation}
so \(\chi^{\mathrm{node}}=3\) and \(\chi^{\mathrm{pair}}=7\).
Likelihood adapters receive the observation and, where present, the covariate row; \Cref{app:poisson} lists the descriptors of the Poisson adapter as a second example.

\paragraph{Shared network.}
The shared network \(\mathtt{net}\) is a stack of four node-pair layers with the inputs of \Cref{eq:np-inputs}, applied to every factor with the same weights.
Its output is the initial embedding \(E_n^{(0)}\) of \Cref{sec:architecture}.

\subsection{Merge}
\label{app:merge}

The merge applies \(M=4\) Transformer blocks to the factor sequence \(E^{(0)}_{0:N}\) and sums the result over factors.
Inside a block, \(b_Q,b_K,b_V,b_{\mathrm{out}},b_{\mathrm{FFN}}\) are node-pair maps, and \(\operatorname{LN}\) is layer normalization over the \(C\) features at each node or pair.
\Cref{tab:boxtransformer_comparison} places the block side by side with a vanilla Transformer layer.

\begin{table}[!htbp]
\centering
\caption{
    Vanilla Transformer versus AFIN merge block for a sequence of length \(N+1\).
    Both compute \(A\in\mathbb R^{(N+1)\times(N+1)}\), but only the vanilla layer uses learned maps whose shapes involve \(DC\).
    The table shows the single-stream notation; the AFIN implementation uses the multi-head factor-axis version described in \Cref{app:merge}, and \(U\) denotes the attention width, equal to \(C\) in AFIN.
}
\label{tab:boxtransformer_comparison}
\small
\setlength{\tabcolsep}{2.5pt}
\renewcommand{\arraystretch}{1.25}
\begin{tabular}{@{}lcc@{}}
\toprule
& Vanilla Transformer & AFIN merge block \\

\midrule
\makecell[l]{Token}
&
\(E_n^{\mathrm{flat}}=\operatorname{vec}(E_n)\in\mathbb R^{DC}\)
&
\(E_n\in\mathbb R^{D\times C}\)
\\
\lightrowrule

\makecell[l]{QKV}
&
\(\begin{gathered}
Q_n=W_Q\operatorname{LN}(E_n^{\mathrm{flat}}),\;
K_n=W_K\operatorname{LN}(E_n^{\mathrm{flat}}),\\
V_n=W_V\operatorname{LN}(E_n^{\mathrm{flat}}),\;
Q_n,K_n,V_n\in\mathbb R^U
\end{gathered}\)
&
\(\begin{gathered}
Q_n=b_Q(\operatorname{LN}(E_n)),\;
K_n=b_K(\operatorname{LN}(E_n)),\\
V_n=b_V(\operatorname{LN}(E_n)),\;
Q_n,K_n,V_n\in\mathbb R^{D\times U}
\end{gathered}\)
\\
\lightrowrule

\makecell[l]{Score\\attention}
&
\(\begin{gathered}
S_{n\ell}=U^{-1/2}\langle Q_n,K_\ell\rangle,\\
A_{n\ell}=\operatorname{softmax}_{\ell}(S_{n\ell}),\;
A\in\mathbb R^{(N+1)\times(N+1)}
\end{gathered}\)
&
\(\begin{gathered}
S_{n\ell}=(D\sqrt U)^{-1}
\sum_{i=1}^D
\langle Q_{n,i},K_{\ell,i}\rangle,\\
A_{n\ell}=\operatorname{softmax}_{\ell}(S_{n\ell}),\;
A\in\mathbb R^{(N+1)\times(N+1)}
\end{gathered}\)
\\
\lightrowrule

\makecell[l]{Value\\output}
&
\(\begin{gathered}
\widetilde V_n=\sum_{\ell=0}^N A_{n\ell}V_\ell\in\mathbb R^U,\\
\widetilde{E}_n^{\mathrm{flat}}
=
E_n^{\mathrm{flat}}
+
W_O\widetilde V_n
\in\mathbb R^{DC}
\end{gathered}\)
&
\(\begin{gathered}
\widetilde{V}_{n,i}
=
\sum_{\ell=0}^N
A_{n\ell}V_{\ell,i},
\;
\widetilde{V}_n\in\mathbb R^{D\times U},\\
\widetilde{E}_n
=
E_n
+
b_{\mathrm{out}}(\widetilde{V}_n)
\in\mathbb R^{D\times C}
\end{gathered}\)
\\
\lightrowrule

\makecell[l]{FFN}
&
\(\begin{gathered}
(E_n^{\mathrm{flat}})^+
=
\widetilde{E}_n^{\mathrm{flat}}
+
\operatorname{FFN}
\left(
\operatorname{LN}(\widetilde{E}_n^{\mathrm{flat}})
\right)
\end{gathered}\)
&
\(\begin{gathered}
E_n^+
=
\widetilde{E}_n
+
b_{\mathrm{FFN}}
\left(
\operatorname{LN}(\widetilde{E}_n)
\right)
\end{gathered}\)
\\
\lightrowrule

\makecell[l]{Maps}
&
\(\begin{gathered}
W_Q,W_K,W_V\in\mathbb R^{U\times DC},\;
W_O\in\mathbb R^{DC\times U}\\
\operatorname{FFN}:\mathbb R^{DC}\to\mathbb R^{DC}
\end{gathered}\)
&
\(\begin{gathered}
b_Q,b_K,b_V:\mathbb R^{D\times C}\to\mathbb R^{D\times U}\\
b_{\mathrm{out}}:\mathbb R^{D\times U}\to\mathbb R^{D\times C},\;
b_{\mathrm{FFN}}:\mathbb R^{D\times C}\to\mathbb R^{D\times C}
\end{gathered}\)
\\
\bottomrule
\end{tabular}

\end{table}

\paragraph{Attention.}
Block \(m\) forms queries, keys, and values from the normalized input, \(Q_{0:N}=b_Q(\operatorname{LN}(E^{(m-1)}_{0:N}))\) and likewise \(K_{0:N}\) and \(V_{0:N}\).
Each is a node-pair object with \(C\) features, split into \(\mathsf h=8\) heads of width \(u=5\); the coordinate axes are never split.
Within a head, the score between factors \(n\) and \(\ell\) averages inner products over coordinates and over coordinate pairs,
\begin{equation}
    S_{n\ell}^{\mathrm{node}}=\frac{1}{D}\sum_{i}\bigl\langle Q_{n,i}^{\mathrm{node}},K_{\ell,i}^{\mathrm{node}}\bigr\rangle,
    \qquad
    S_{n\ell}^{\mathrm{pair}}=\frac{1}{D^2}\sum_{i,j}\bigl\langle Q_{n,ij}^{\mathrm{pair}},K_{\ell,ij}^{\mathrm{pair}}\bigr\rangle,
\label{eq:merge-scores}
\end{equation}
and the attention weights are
\begin{equation}
    A_{n\ell}=\operatorname{softmax}_{\ell}\Bigl(\frac{\lambda_{\mathrm{node}}S_{n\ell}^{\mathrm{node}}+\lambda_{\mathrm{pair}}S_{n\ell}^{\mathrm{pair}}}{\sqrt u}\Bigr),
    \qquad n,\ell=0,\ldots,N,
\end{equation}
with learned mixing weights \(\lambda_{\mathrm{node}}+\lambda_{\mathrm{pair}}=1\) per block.
The same weights average the node and pair values of all factors, coordinate by coordinate,
\begin{equation}
    \widetilde V_{n,i}^{\mathrm{node}}=\sum_{\ell=0}^{N}A_{n\ell}V_{\ell,i}^{\mathrm{node}},
    \qquad
    \widetilde V_{n,ij}^{\mathrm{pair}}=\sum_{\ell=0}^{N}A_{n\ell}V_{\ell,ij}^{\mathrm{pair}},
\end{equation}
and the heads are concatenated along the feature axis.

\paragraph{Residual updates.}
The attention output and the feed-forward update are added as residuals, with the pair array symmetrized each time,
\begin{equation}
\begin{aligned}
    \widetilde E_n &= \operatorname{sym}\bigl(E_n^{(m-1)}+b_{\mathrm{out}}(\widetilde V_n)\bigr),\\
    E_n^{(m)} &= \operatorname{sym}\bigl(\widetilde E_n+b_{\mathrm{FFN}}(\operatorname{LN}(\widetilde E_n))\bigr),
\end{aligned}
\end{equation}
where \(\operatorname{sym}\) acts on the pair component only.
No positional encodings are used, so the block is equivariant to the order of the factors.

\paragraph{Pooling.}
After the last block, \(\bar E=\sum_{n=0}^{N}E_n^{(M)}\), symmetrized once more.
Pooling has no parameters, so the merge accepts any \(N\).

\subsection{Decoder}
\label{app:decode}

Three coordinate-wise MLPs with two hidden layers of width \(160\) map \(\bar E\) to the Gaussian parameters.
The mean MLP takes a \(5C\)-dimensional input at each coordinate, namely the node features, the average of the pair row and column means, the diagonal pair features, and the global node and pair averages, and outputs \(\mu_i\).
The off-diagonal MLP maps each pair's features to a scalar; these are scaled by \(0.1\), symmetrized, and placed in a matrix \(\Omega\) with zero diagonal.
The diagonal MLP maps each node's features to a scalar \(r_i\).
The precision is then
\begin{equation}
    (\Lambda_\phi)_{ij}=
    \begin{cases}
        \Omega_{ij}, & i\ne j,\\
        \sum_{k\ne i}|\Omega_{ik}|+\operatorname{softplus}(r_i)+10^{-4}, & i=j,
    \end{cases}
\end{equation}
which is symmetric and strictly diagonally dominant with a positive diagonal, hence positive definite.
Its Cholesky factor is used for density evaluation and sampling.

\subsection{Configuration and parameter count}
\label{app:model-config}

The default configuration uses \(C=40\) features; coordinate-wise MLPs with two hidden layers of width \(192\) in the encoder and merge and of width \(160\) in the decoder; adapters with one hidden layer of width \(64\); four node-pair layers in the shared encoder; and \(M=4\) merge blocks with \(8\) heads.
\Cref{tab:afin-params} gives the parameter count by module; none of the counts depends on \(D\) or \(N\).

\begin{table}[!htbp]
\centering
\caption{
    Module-wise parameter counts for the default AFIN configuration.
    Counts are independent of the latent dimension \(D\) and the number of likelihood factors \(N\).
}
\label{tab:afin-params}
\begin{tabular}{lr}
\toprule
Module & \# parameters \\
\midrule
\quad Factor-type adapters        & 55{,}376 \\
\quad Shared encoder (four node-pair layers)              & 728{,}384 \\
\quad Merge blocks (\(M=4\)) & 3{,}643{,}208 \\
\quad Gaussian decoder            & 123{,}043 \\
\midrule
Total trainable parameters & 4{,}550{,}011 \\
\bottomrule
\end{tabular}
\end{table}

\subsection{Computational cost}
\label{app:complexity}

Let \(K=N+1\) be the number of factors.
Every map is applied at \(D\) nodes and \(D^2\) pairs per factor, so the encoder and the five maps inside each merge block cost \(O(KD^2(CH+H^2))\) operations and hold \(O(KD^2C)\) activations per layer.
Attention adds \(O(K^2D^2C)\) for the scores in \Cref{eq:merge-scores} and the same again for averaging the values.
The decoder costs \(O(D^2(CH+H^2))\) plus \(O(D^3)\) for the Cholesky factorization, and each SNIS proposal costs one Gaussian sample and one evaluation of the model's log density.
The pair arrays therefore make the cost quadratic in \(D\), and attention makes it quadratic in \(N\); the learned parameters in \Cref{tab:afin-params} are unaffected, and \Cref{app:scaling} reports measured times and memory.

\section{Training Details}
\label{app:simulator}

We train on tasks generated on the fly, so no task is ever reused.
Each optimizer step uses four micro-batches of \(B=32\) tasks.
Tasks in a micro-batch share \(D\) and \(N\) so that they can be batched; everything else is sampled independently for each task.

\paragraph{Task sizes.}
With probability \(0.6\) we draw \((D,N)\) from
\begin{equation}
    \mathbb P(D,N)\propto\Bigl(\frac{D}{16}\Bigr)^{1.0}\Bigl(\frac{1}{N}\Bigr)^{0.75},
    \qquad D\in\{1,\ldots,16\},\; N\in\{1,\ldots,256\},
\end{equation}
and otherwise uniformly from the same ranges.
The bias puts more weight on high-dimensional, data-poor tasks, which are the hard ones under uniform sampling.

\paragraph{Prior.}
One of four configurations, chosen uniformly.
The location is \(\mu\sim0.45\,\mathcal N(0,I_D)\) in all cases, and the remaining parameters are
\begin{itemize}[leftmargin=2em,topsep=0.25em,itemsep=0.1em]
    \item \texttt{gaussian prior}, diagonal covariance: \(\log\sigma_i\sim\mathrm{Uniform}(-0.8,0)\);
    \item \texttt{gaussian prior}, full covariance: precision \(\Lambda=GG^\top/D+0.5I_D\) with \(G\sim0.3\,\mathcal N(0,I_{D\times D})\);
    \item \texttt{student-t prior}: \(\log\sigma_i\sim\mathrm{Uniform}(-0.7,0)\) and \(\nu\sim\mathrm{Uniform}(3,8)\);
    \item \texttt{laplace prior}: \(\log s_i\sim\mathrm{Uniform}(-1.0,-0.05)\).
\end{itemize}
We then draw \(z\) from this prior.

\paragraph{Likelihoods.}
There are five likelihood forms: \texttt{gaussian likelihood} with or without covariates, \texttt{student-t likelihood} with covariates, \texttt{bernoulli-logit likelihood}, and \texttt{binomial-logit likelihood}.
For \(N=1\) we pick one form uniformly.
For \(N>1\), with probability \(0.5\) all observations share one uniformly chosen form; otherwise we draw the number of distinct forms \(k\in\{2,\ldots,\min(5,N)\}\) with \(\mathbb P(k)\propto e^{-(k-2)}\), pick \(k\) forms without replacement, allocate observations to them with weights \(\pi\sim\mathrm{Dirichlet}(0.5\,\mathbf 1_k)\) and at least one observation per form, and permute the assignment.
With covariates, \(y_n=x_n^\top z+\varepsilon_n\) with Gaussian or Student-\(t\) noise of sampled scale, \(y_n\sim\mathrm{Bernoulli}(\sigma(x_n^\top z))\), or \(y_n\sim\mathrm{Binomial}(n_c,\sigma(x_n^\top z))\) with \(n_c\sim\mathrm{Uniform}\{2,\ldots,8\}\).
Without covariates, \(y_n=z+\varepsilon_n\in\mathbb R^D\) with isotropic Gaussian noise of sampled scale.

\paragraph{Covariates.}
The design matrix \(X\in\mathbb R^{N\times D}\) has base scale \(0.9/\sqrt D\), which keeps \(Xz\) of order one.
Its rows follow one of four distributions with probabilities \((0.7,0.1,0.1,0.1)\): isotropic Gaussian; Gaussian with a sampled diagonal spectrum; Gaussian with a sampled spectrum and a random rotation; and the latter with row-wise multiplicative weights that give multivariate Student-\(t\) tails.

\paragraph{Optimization.}
AdamW for \(100{,}000\) steps with cosine learning-rate decay from \(2\times10^{-4}\) and gradient accumulation over the four micro-batches; evaluation uses an exponential moving average of the weights.
Training takes under 24 hours on one H100 GPU.

\section{Evaluation Details}
\label{app:evaluation-protocol}

\subsection{Baselines and timing}
\label{app:baselines}

The following protocol applies to \Cref{fig:synth-posterior-time,fig:uci-results,fig:synth-frvi-snis} and their supplementary figures.
The fixed-budget extrapolation, Poisson, ablation, and NPE-PFN studies retain the protocols stated in their respective subsections.

\paragraph{NUTS.}
We report both NumPyro and Pyro, using four chains, dense mass adaptation, target acceptance \(0.8\), and maximum tree depth \(10\).
Chains start from independently perturbed prior locations and run for \(1{,}000\) warmup steps, followed by up to \(100{,}000\) retained draws per chain.
Each curve records cumulative time and error at 10, 30, 100, 300, 1k, 3k, 10k, 30k, and 100k retained draws per chain from one continuous run.
All plotted estimates exclude warmup draws, while their times include warmup and, for NumPyro, model-specific JAX compilation.

\paragraph{Variational baselines.}
FRVI fits a full-rank Gaussian by stochastic gradient ascent on the ELBO; IAF, MAF, and NSF VI fit flows the same way.
We select one fixed learning rate per method from seven values spaced logarithmically between \(10^{-6}\) and \(10^{-2}\), using mean negative ELBO per dimension on 16 separate validation tasks after \(3{,}000\) steps.
The selected rates are approximately \(2.15\times10^{-3}\) for FRVI and NSF, \(10^{-4}\) for IAF, and \(10^{-2}\) for MAF; benchmark references are not used for selection.
FRVI parameterizes its covariance by a Cholesky factor and starts from the prior moments.
All variants use Adam, 16 reparameterized samples per step, and gradient clipping at norm 10.
The flows use four autoregressive transforms with two 64-unit hidden layers, alternating coordinate reversals; NSF uses eight-bin quadratic splines.
FRVI runs for up to 30k steps and the flows for 10k; flow plots omit the 10-step point.
The FRVI proposals for SNIS are snapshots after 1k, 10k, or 30k steps of the same optimization run.

\paragraph{AFIN and SNIS.}
For AFIN, we report the median time over three forward evaluations, including transfer of task inputs to the GPU and construction of the Gaussian posterior.
For each AFIN or FRVI proposal, we run three independent SNIS streams and record cumulative prefixes from 100 to 500 million draws.
Times include proposal construction or the full FRVI fit, target setup, sampling, target and proposal density evaluation, weight normalization, and weighted moments.
All reported SNIS results use the Gaussian proposal directly.

\paragraph{Aggregation.}
For each task and budget, we average errors and take median times over repeated runs, then average tasks equally within each setting.
The homogeneous synthetic settings each contain 16 prior--likelihood combinations and three independently generated tasks per combination; each heterogeneous setting contains 12 combinations and three tasks each.
UCI curves use three VI fits per method and one four-chain NUTS run per backend and dataset.
SNIS shades span the minimum and maximum of three stream-wise task-average errors; they are not confidence intervals or variation across independently trained AFINs.

\paragraph{Hardware and code.}
AFIN and SNIS run on one NVIDIA A100 SXM4 GPU with 80 GB of memory; VI and both NUTS implementations use one pinned AMD EPYC 9474F (Genoa) CPU core.
In preliminary comparisons on ten synthetic tasks, FRVI and NumPyro NUTS were faster on CPU than on an H100 GPU; this informed our choice of CPU execution for the iterative baselines.
Framework imports, checkpoint loading, backend initialization, offline reference construction, reference-based scoring, and file I/O are excluded from the clocks.
VI times include fitting and constructing the approximation; draws used only to score a fitted flow are excluded.
Code is available at \url{https://github.com/joohwanko/AFINs}.

\subsection{Reference posteriors}
\label{app:references}
\label{app:mcmc-diagnostics}

Reference posteriors are computed offline and never enter the inference timings.
For the 144 homogeneous synthetic tasks, we reuse four-chain NumPyro references with \(10{,}000\) warmup steps and \(10^6\) retained draws per chain, target acceptance \(0.8\), maximum tree depth \(10\), and diagonal mass adaptation.
For the 72 heterogeneous and 12 UCI tasks, we run four chains with \(10{,}000\) warmup steps and \(250{,}000\) retained draws per chain, target acceptance \(0.9\), maximum tree depth \(12\), and dense mass adaptation.
Every method is evaluated against the reference for its exact frozen task data and model parameters.
\Cref{tab:mcmc-diagnostics} reports rank-normalized split-\(\widehat R\) and bulk and tail effective sample sizes \citep{vehtari2021rank}.
The homogeneous, heterogeneous, and UCI reference runs have no retained divergences or maximum-depth saturation; their minimum energy Bayesian fractions of missing information are \(0.677\), \(0.709\), and \(1.010\), respectively \citep{betancourt2016diagnosing}.
The 24-task Poisson extension of \Cref{app:poisson} and the 30-task conjugate check below also use four chains with \(10^6\) draws per chain.

\begin{table}[!htbp]
\centering
\caption{Diagnostics of the references used for the reported task sets. Values are the worst over coordinates and tasks; ESS values are rounded to integers.}
\label{tab:mcmc-diagnostics}
\small
\begin{tabular}{lrrrr}
\toprule
Task set & Tasks & Max. $\widehat R$ & Min. bulk ESS & Min. tail ESS \\
\midrule
Homogeneous synthetic & 144 & 1.0000103 & 2,430,159 & 1,463,410 \\
Heterogeneous synthetic & 72 & 1.0000319 & 649,438 & 229,483 \\
UCI & 12 & 1.0000194 & 924,700 & 697,898 \\
Poisson extension & 24 & 1.0000055 & 1,964,364 & 1,567,853 \\
Conjugate check & 30 & 1.0000058 & 3,624,934 & 2,873,407 \\
\bottomrule
\end{tabular}
\end{table}

\paragraph{Check against exact posteriors.}
Thirty conjugate tasks, with diagonal or full-covariance Gaussian priors and linear-Gaussian likelihoods at the three benchmark sizes plus \((D,N)=(16,256)\) and \((32,512)\), three seeds each, have closed-form posteriors.
Against them the long references have mean error \(0.00063\), relative covariance error \(0.0026\), and Gaussian W\(_2\) \(0.0023\), so reference error is negligible at the scale of the results.
The same tasks also test AFIN with no reference in the loop: at the three benchmark sizes (18 tasks) the direct prediction has mean error \(0.047\), relative covariance error \(0.073\), and Gaussian W\(_2\) \(0.054\), with 90\% intervals covering \(0.92\) of the true coordinates, and SNIS brings the errors to \(0.012\), \(0.045\), and \(0.020\) with coverage \(0.89\).

\paragraph{References for the additional experiments.}
The prior--likelihood and merge comparisons in \Cref{app:additional-ablations} reuse the homogeneous synthetic references.
The 128 Poisson-composition tasks use separate references: four dense-metric NumPyro NUTS chains with 2,000 warmup steps and 20,000 retained draws per chain.
All new runs satisfy rank-normalized split-\(\widehat R\le1.01\), bulk ESS at least 1,000, and tail ESS at least 400.
They have no divergences, E-BFMI at least 0.3, and maximum-depth saturation below 1\%.
For these additional experiments, reference moments are computed from the last two chains; SW\(_2\) and joint posterior mass use 16,384 draws from those chains.

\subsection{Metrics}
\label{app:metrics}
\label{app:evaluation-metrics}

We measure posterior mean and covariance errors by
\begin{equation}
    \mathrm{M1}=\bigl\|\widehat\mu_q-\widehat\mu_{\mathrm{ref}}\bigr\|_2,
    \qquad
    \mathrm{M2}=\bigl\|\widehat\Sigma_q-\widehat\Sigma_{\mathrm{ref}}\bigr\|_F.
\end{equation}
For AFIN and FRVI, the approximation's moments are available in closed form; for NUTS and flows they are empirical moments.
SNIS uses normalized weights \(\bar\omega_s\propto p(z_q^{(s)},y)/q(z_q^{(s)})\) to compute weighted moments.
For the time--error curves, reference moments use all retained reference draws.

Sliced Wasserstein-2 compares one-dimensional projections along \(R=128\) random unit directions shared by all methods for a task \citep{bonneel2015sliced}.
For the time--error curves, we evaluate the projected quantile functions on \(K=32{,}768\) midpoint probabilities \(u_k=(k-1/2)/K\):
\begin{equation}
    \mathrm{SW}_2=\left[\frac{1}{RK}\sum_{r=1}^{R}\sum_{k=1}^{K}
    \bigl(Q_{q,r}(u_k)-Q_{\mathrm{ref},r}(u_k)\bigr)^2\right]^{1/2}.
\end{equation}
Here \(Q_{q,r}\) is the Gaussian quantile function for AFIN and FRVI, the empirical projected quantile function for NUTS and flows, and its weighted empirical version for SNIS.
Reference quantiles use \(262{,}144\) draws balanced across chains; flow quantiles use \(65{,}536\) samples.
At each SNIS budget, every proposal draw contributes to the weighted empirical distribution.
Moment accumulation uses float64, and reference scoring is excluded from the reported inference time.
The original fixed-budget studies use sample-based SW\(_2\) estimates.
The new training comparisons in \Cref{app:additional-ablations} use Gaussian projected quantiles with 64 directions and 16,384 reference draws per task.

\paragraph{Coverage.}
On simulated tasks the true latent vector is known, so we also check calibration: the fraction of latent coordinates whose true value lies inside the approximation's central marginal interval of nominal level \(1-\alpha\), computed from the Gaussian quantiles for the direct prediction and from weighted quantiles of the proposals for SNIS, and averaged over coordinates, tasks, and seeds.
The ideal value is \(1-\alpha\); smaller values mean the posterior is overconfident.

\paragraph{Poisson composition metrics.}
For \Cref{tab:poisson-composition}, we normalize each coordinate's mean error by its reference posterior standard deviation, then take the root mean square across coordinates.
Relative covariance error is the Frobenius error divided by the reference covariance's Frobenius norm.
Joint posterior mass is the fraction of reference draws inside the predicted Gaussian's nominal 90\% ellipsoid; it measures posterior mass within a task.

\paragraph{Weight diagnostics.}
We report the importance-weight ESS fraction \(1/(S\sum_s\bar\omega_s^2)\), the maximum normalized weight \(\max_s\bar\omega_s\), and the estimated Pareto-\(k\) of the upper weight tail \citep{vehtari2024psis}.
Higher ESS fractions and lower maximum weights indicate less concentrated weights; larger \(k\) indicates a heavier tail.
Weights are not Pareto-smoothed in our estimates.
These finite-sample diagnostics do not guarantee finite weight variance.

\subsection{Synthetic benchmark}
\label{app:synthetic-tasks}

The homogeneous suite crosses the four prior configurations of \Cref{app:simulator} with the four likelihood forms that use covariates, giving 16 combinations at each of three sizes, with three independent task seeds per combination (144 tasks).
The Gaussian likelihood without covariates is trained on but not included in this suite.
We use the descriptive setting names in \Cref{tab:synthetic-regimes}, rather than treating size alone as a difficulty ranking.

The heterogeneous suite combines each prior configuration with three likelihood mixtures: Gaussian and Bernoulli-logit; Student-\(t\) and Bernoulli-logit; or all three.
For each mixture, observations are allocated as evenly as possible across its types and the assignments are permuted.
We use \(D=16\), \(N\in\{8,32\}\), and three task seeds per combination (72 tasks), with correlated Gaussian covariates whose population covariance has condition number up to 40.
The remaining factor parameters follow the training simulator.
These tasks test inference with several supported likelihood types in one model; mixtures are also present during training, so this is not a held-out-combination test.
AFIN receives only the factor specification and the observations.

\begin{table}[!htbp]
\centering
\caption{Synthetic evaluation settings. The heterogeneous tasks mix likelihood types; the other settings use one likelihood type per task.}
\label{tab:synthetic-regimes}
\begin{tabular}{lrrr}
\toprule
Setting & Latent dimension \(D\) & Observations \(N\) & Tasks \\
\midrule
Data-rich & 4 & 256 & 48 \\
Intermediate & 8 & 64 & 48 \\
Data-scarce & 16 & 1 & 48 \\
Heterogeneous & 16 & 8 & 36 \\
Heterogeneous (appendix) & 16 & 32 & 36 \\
\bottomrule
\end{tabular}
\end{table}

\subsection{UCI tasks}
\label{app:uci-tasks}

\Cref{tab:uci-task-specs} lists the 12 tasks.
Each is a Bayesian regression or classification model with one prior over the coefficient vector \(z\) and one likelihood factor per row; continuous covariates are standardized and whitened, and regression targets are standardized.
After whitening, covariates are scaled by \(0.9/\sqrt P\), where \(P\) is the number of selected covariates; Mesquite additionally has an unscaled intercept.
Whitening caps the inverse square roots of covariance eigenvalues at 10.
Prior locations are zero and scales are one, Student-\(t\) priors and likelihoods have four degrees of freedom, and continuous likelihoods have fixed noise scale \(0.5\).
We use the full datasets listed in the table and freeze preprocessing, observations, and factor parameters before running any method.
The two heterogeneous tasks start from a regression dataset, keep half of the rows as continuous observations, and turn the other half into binary labels that say whether the response is above the dataset median.
Forest Fires models \(\log(1+\mathrm{area})\) with a Student-\(t\) likelihood and Mesquite models biomass with a Gaussian likelihood; the labels use Bernoulli-logit likelihoods in both, and both observation types share one coefficient vector.
The Bernoulli-logit factor is a modeling choice for these labels, not the likelihood implied by thresholding the continuous model.

\begin{table}[!htbp]
\centering
\caption{
    Real-world task specifications, using the type labels from \Cref{sec:experimental-setup}.
    All priors have independent coordinates, and Gaussian and Student-t likelihoods use covariates.
    Rows listing two likelihood types use both within one model.
    Dataset sources are cited below.
    \citep{concrete_slump_test_182,efron2004least,yacht_hydrodynamics_243,computer_hardware_29,auto_mpg_9,habermans_survival_43,connectionist_bench_sonar,parkinsons_174,heart_disease_45,seeds_236,forest_fires_162,Gelman_Hill_2006}
}
\label{tab:uci-task-specs}
\begin{tabular}{@{}lrrll@{}}
\toprule
Dataset & \(D\) & \(N\) & Prior type & Likelihood type(s) \\
\midrule
Concrete Slump          & \(7\)  & \(103\) & \texttt{laplace prior}   & \texttt{gaussian likelihood} \\
Diabetes                & \(10\) & \(442\) & \texttt{laplace prior}   & \texttt{gaussian likelihood} \\
Yacht                   & \(6\)  & \(308\) & \texttt{gaussian prior}  & \texttt{student-t likelihood} \\
Machine CPU             & \(6\)  & \(209\) & \texttt{laplace prior}   & \texttt{gaussian likelihood} \\
Auto MPG                & \(7\)  & \(392\) & \texttt{laplace prior}   & \texttt{gaussian likelihood} \\
Haberman                & \(3\)  & \(306\) & \texttt{gaussian prior}  & \texttt{bernoulli-logit likelihood} \\
Sonar top-8             & \(8\)  & \(208\) & \texttt{student-t prior} & \texttt{bernoulli-logit likelihood} \\
\makecell[l]{Parkinsons Voice\\top-8} & \(8\) & \(195\) & \texttt{gaussian prior} & \texttt{bernoulli-logit likelihood} \\
Heart                   & \(13\) & \(297\) & \texttt{gaussian prior}  & \texttt{bernoulli-logit likelihood} \\
Seeds Binary            & \(7\)  & \(210\) & \texttt{gaussian prior}  & \texttt{bernoulli-logit likelihood} \\
Forest Fires Hetero     & \(8\)  & \(517\) & \texttt{student-t prior} & \makecell[l]{\texttt{student-t likelihood}\\\texttt{bernoulli-logit likelihood}} \\
Mesquite Hetero         & \(7\)  & \(46\)  & \texttt{gaussian prior}  & \makecell[l]{\texttt{gaussian likelihood}\\\texttt{bernoulli-logit likelihood}} \\
\bottomrule
\end{tabular}

\end{table}

\clearpage
\subsection{Extrapolation}
\label{app:extrapolation}

The six out-of-range configurations are \((D,N)=(24,128)\) and \((32,64)\) for OOD \(D\); \((12,400)\) and \((8,512)\) for OOD \(N\); and \((24,384)\) and \((32,400)\) for OOD \(D,N\), each with three seeds.
This is a separate fixed-budget evaluation on an A100 GPU, with Pyro NUTS and per-task learning-rate selection for FRVI.
AFIN draws \(5\times10^4\) samples, AFIN+SNIS uses \(5\times10^4\) proposals, NUTS draws \(5\times10^4\) samples after \(10^3\) warmup steps, and FRVI is optimized for \(10^4\) steps at the best of seven learning rates and then sampled \(5\times10^4\) times.
\Cref{tab:extrapolate-highdn-sw2,tab:extrapolate-highdn-m1,tab:extrapolate-highdn-m2} give means and standard deviations over the six runs in each split, with the mean time per task.

\begin{table}[H]
\centering
\caption{Extrapolation results for SW$_2$.
    Each out-of-distribution (OOD) split contains two configurations and three seeds.
    SW$_2$ is mean $\pm$ standard deviation; time is the mean per task. Best in bold.}
\label{tab:extrapolate-highdn-sw2}
\small
\setlength{\tabcolsep}{5pt}
\begin{tabular}{@{}lrrrrrr@{}}
\toprule
& \multicolumn{2}{c}{OOD \(D\)}
& \multicolumn{2}{c}{OOD \(N\)}
& \multicolumn{2}{c}{OOD \(D,N\)} \\
\cmidrule(lr){2-3}\cmidrule(lr){4-5}\cmidrule(l){6-7}
Method & SW$_2$ & Time (s) & SW$_2$ & Time (s) & SW$_2$ & Time (s) \\
\midrule
AFIN
& $0.031\pm0.010$ & 0.02
& $0.027\pm0.004$ & 0.02
& $0.040\pm0.020$ & 0.10 \\
AFIN+SNIS
& $0.024\pm0.020$ & 0.24
& $\mathbf{0.0013}\pm0.0001$ & 0.25
& $\mathbf{0.0039}\pm0.0010$ & 0.45 \\
NUTS
& $\mathbf{0.012}\pm0.006$ & 308
& $0.0025\pm0.0003$ & 95
& $0.0052\pm0.0008$ & 157 \\
FRVI
& $0.042\pm0.040$ & 27
& $0.0025\pm0.0004$ & 32
& $0.0041\pm0.0007$ & 100 \\
\bottomrule
\end{tabular}
\end{table}

\begin{table}[H]
\centering
\caption{
    Extrapolation results for posterior mean error.
    Entries report mean \(L_2\) error, mean \(\pm\) standard deviation over six runs, with mean wall-clock time in parentheses. Best in bold.}
\label{tab:extrapolate-highdn-m1}
\setlength{\tabcolsep}{4pt}
\begin{tabular}{lcccc}
\toprule
Split & AFIN & AFIN+SNIS & NUTS & FRVI \\
\midrule
OOD \(D\)
& \makecell{\(0.134\pm0.044\)\\(0.02s)}
& \makecell{\(0.068\pm0.063\)\\(0.24s)}
& \makecell{\(\mathbf{0.0277}\pm0.019\)\\(308s)}
& \makecell{\(0.0872\pm0.074\)\\(27s)} \\
OOD \(N\)
& \makecell{\(0.0799\pm0.011\)\\(0.02s)}
& \makecell{\(\mathbf{0.00185}\pm0.0006\)\\(0.25s)}
& \makecell{\(0.00248\pm0.0008\)\\(95s)}
& \makecell{\(0.00603\pm0.002\)\\(32s)} \\
OOD \(D,N\)
& \makecell{\(0.205\pm0.103\)\\(0.10s)}
& \makecell{\(\mathbf{0.00984}\pm0.005\)\\(0.45s)}
& \makecell{\(0.0110\pm0.004\)\\(157s)}
& \makecell{\(0.0143\pm0.003\)\\(100s)} \\
\bottomrule
\end{tabular}

\end{table}

\begin{table}[H]
\centering
\caption{
    Extrapolation results for posterior covariance error.
    Entries report covariance Frobenius error, mean \(\pm\) standard deviation over six runs, with mean wall-clock time in parentheses. Best in bold.}
\label{tab:extrapolate-highdn-m2}
\setlength{\tabcolsep}{4pt}
\begin{tabular}{lcccc}
\toprule
Split & AFIN & AFIN+SNIS & NUTS & FRVI \\
\midrule
OOD \(D\)
& \makecell{\(0.234\pm0.164\)\\(0.02s)}
& \makecell{\(0.323\pm0.307\)\\(0.24s)}
& \makecell{\(\mathbf{0.131}\pm0.105\)\\(308s)}
& \makecell{\(0.384\pm0.356\)\\(27s)} \\
OOD \(N\)
& \makecell{\(0.00580\pm0.0023\)\\(0.02s)}
& \makecell{\(\mathbf{0.00115}\pm0.0004\)\\(0.25s)}
& \makecell{\(0.00217\pm0.0009\)\\(95s)}
& \makecell{\(0.00292\pm0.0009\)\\(32s)} \\
OOD \(D,N\)
& \makecell{\(0.0538\pm0.027\)\\(0.10s)}
& \makecell{\(\mathbf{0.0175}\pm0.008\)\\(0.45s)}
& \makecell{\(0.0222\pm0.007\)\\(157s)}
& \makecell{\(0.0248\pm0.006\)\\(100s)} \\
\bottomrule
\end{tabular}

\end{table}

\clearpage
\section{Additional Results}
\label{app:additional-results}

\subsection{Additional synthetic results}

\begin{figure}[H]
    \centering
    \includegraphics[width=\linewidth]{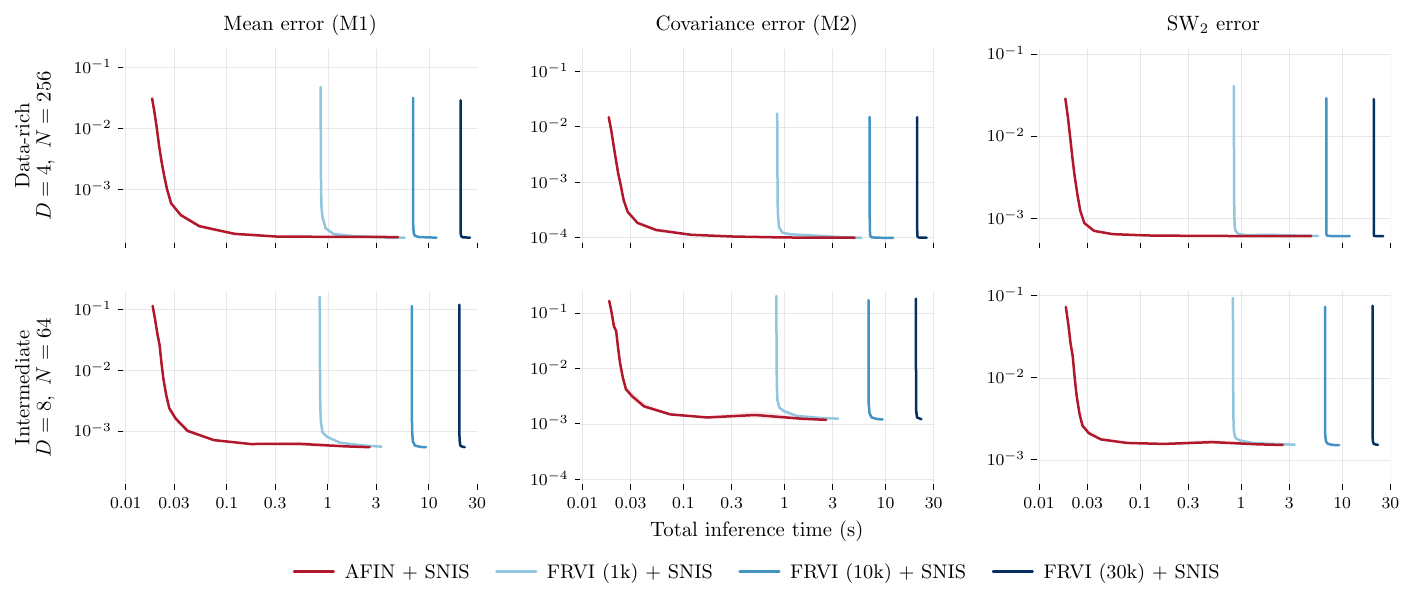}
    \caption{SNIS refinement on data-rich \((D=4,N=256)\) and intermediate \((D=8,N=64)\) tasks, with 48 tasks per row. Same proposals, budgets, and timing as \Cref{fig:synth-frvi-snis}.}
    \label{fig:snis-supporting}
\end{figure}

\begin{figure}[H]
    \centering
    \includegraphics[width=\linewidth]{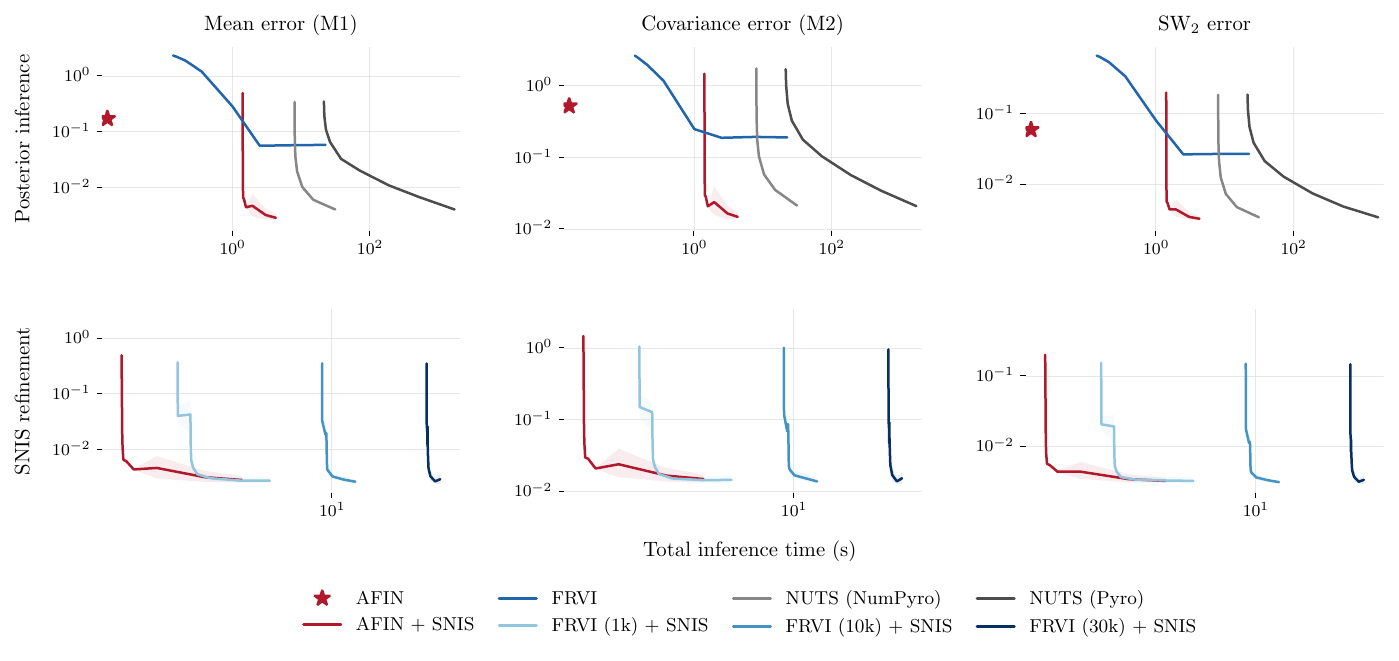}
    \caption{Heterogeneous likelihoods at \(D=16,N=32\), averaged over 36 tasks. Top: direct AFIN, AFIN+SNIS, FRVI, and both NUTS implementations. Bottom: AFIN and FRVI proposals under the same SNIS budgets, including proposal construction or optimization time.}
    \label{fig:heterogeneous-n32}
\end{figure}

\Cref{fig:snis-supporting} complements \Cref{fig:synth-frvi-snis}: on data-rich and intermediate tasks, AFIN and the longer FRVI fits reach similar errors after SNIS, while AFIN avoids per-task optimization.
\Cref{fig:heterogeneous-n32} reports the 36 heterogeneous tasks at \(N=32\), using the same likelihood mixtures and inference budgets as the \(N=8\) setting.

\clearpage
\subsection{UCI mean and covariance errors}

\begin{figure}[H]
    \centering
    \includegraphics[width=1.0\linewidth]{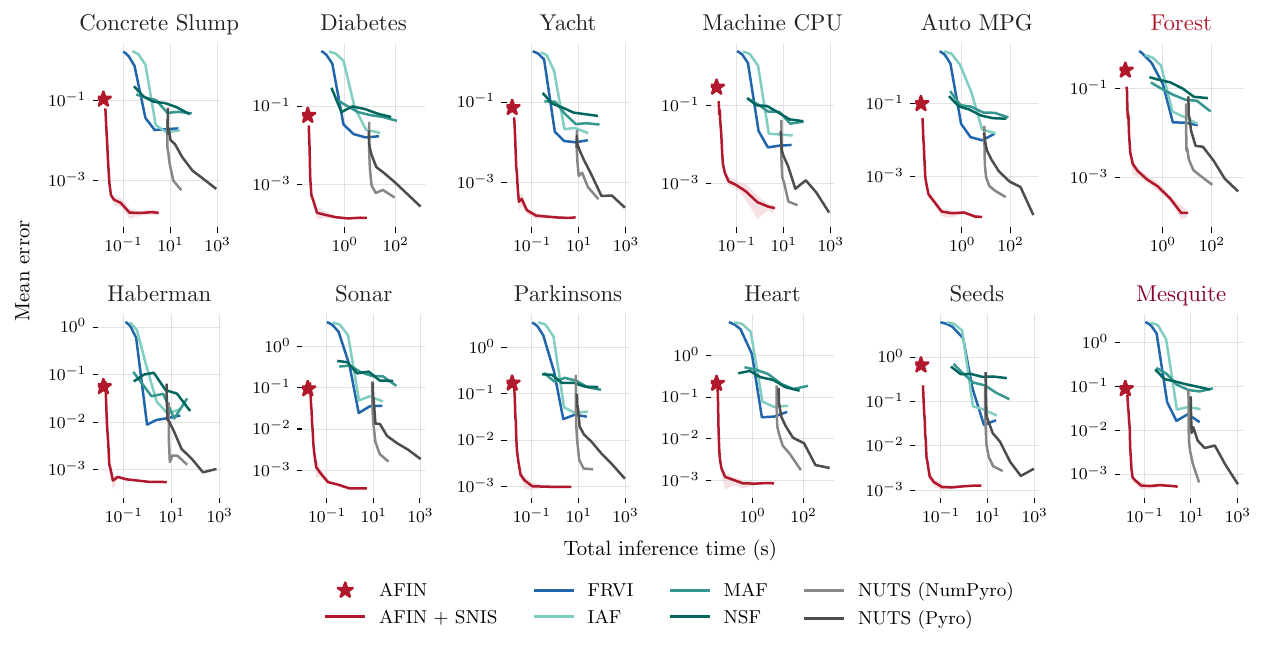}
    \caption{Posterior mean error (M1) against inference time on the 12 UCI tasks; same protocol as \Cref{fig:uci-results}.}
    \label{fig:uci-m1-results}
\end{figure}

\begin{figure}[H]
    \centering
    \includegraphics[width=1.0\linewidth]{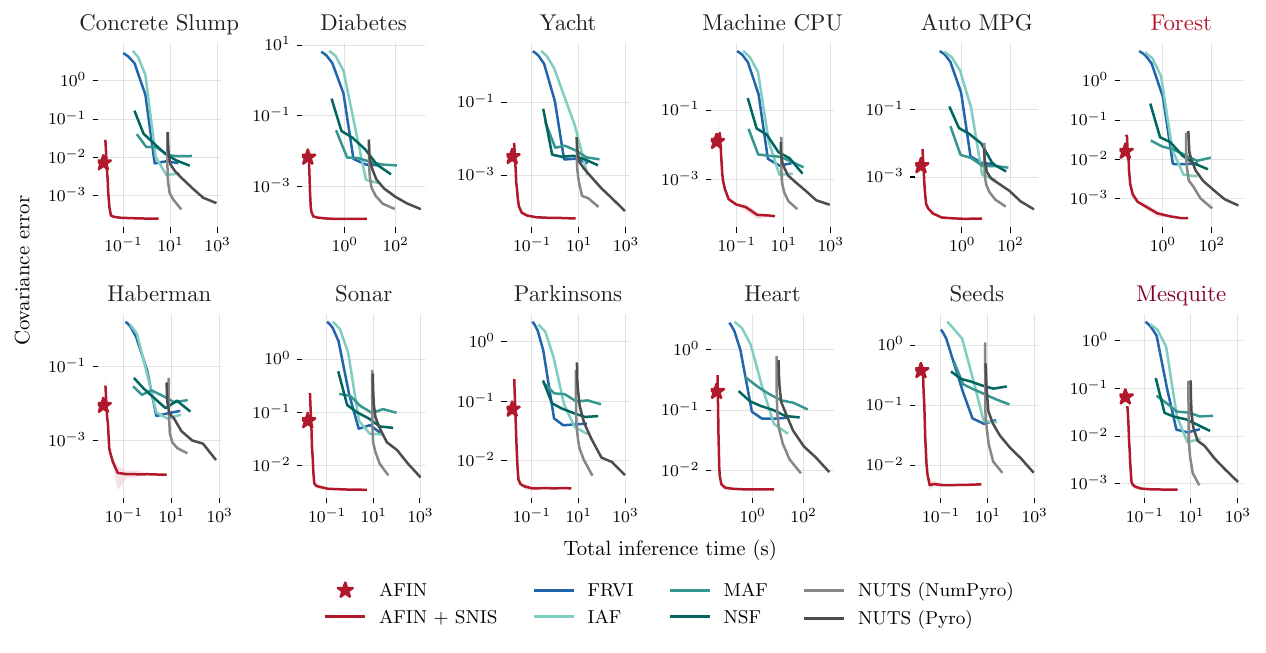}
    \caption{Posterior covariance error (M2) against inference time on the 12 UCI tasks; same protocol as \Cref{fig:uci-results}.}
    \label{fig:uci-m2-results}
\end{figure}

\Cref{fig:uci-m1-results,fig:uci-m2-results} report M1 and M2 for the UCI tasks, complementing the SW\(_2\) curves in \Cref{fig:uci-results}.
SNIS substantially reduces the mean and covariance errors of the direct AFIN predictions; the plots show how these gains compare with longer VI and NUTS runs.

\clearpage
\subsection{SNIS weight diagnostics}
\label{app:snis-stability}

\Cref{tab:snis-diagnostics} compares the same four proposals as \Cref{fig:synth-frvi-snis} at 500 million SNIS draws.
We first average each diagnostic over three sampling streams within a task, then average tasks within each setting.
The final column reports the percentage of tasks whose stream-averaged Pareto-\(k\) exceeds \(0.7\).

\begin{table}[!htbp]
\centering
\small
\setlength{\tabcolsep}{4pt}
\caption{SNIS weight diagnostics at 500 million proposals. Entries are task means after averaging three sampling streams; the final column is the percentage of tasks with estimated \(k>0.7\). Higher ESS fractions and lower maximum weights indicate less concentrated weights.}
\label{tab:snis-diagnostics}
\begin{tabular}{@{}llrrrr@{}}
\toprule
Setting & Proposal & ESS / samples & Max. weight & Pareto-$k$ & $k>0.7$ (\%) \\
\midrule
Data-rich & AFIN & 0.954 & $6.77\times10^{-8}$ & 0.155 & 2.1 \\
 & FRVI (1k) & 0.397 & $6.18\times10^{-7}$ & -0.354 & 2.1 \\
 & FRVI (10k) & 0.971 & $9.14\times10^{-8}$ & 0.172 & 0.0 \\
 & FRVI (30k) & 0.976 & $8.48\times10^{-8}$ & 0.174 & 0.0 \\
\midrule
Intermediate & AFIN & 0.836 & $5.85\times10^{-6}$ & 0.274 & 0.0 \\
 & FRVI (1k) & 0.532 & $1.09\times10^{-5}$ & 0.195 & 0.0 \\
 & FRVI (10k) & 0.842 & $9.31\times10^{-6}$ & 0.303 & 0.0 \\
 & FRVI (30k) & 0.839 & $8.92\times10^{-6}$ & 0.303 & 0.0 \\
\midrule
Data-scarce & AFIN & 0.473 & $7.29\times10^{-4}$ & 0.444 & 25.0 \\
 & FRVI (1k) & 0.477 & $1.50\times10^{-3}$ & 0.476 & 47.9 \\
 & FRVI (10k) & 0.477 & $1.14\times10^{-3}$ & 0.475 & 50.0 \\
 & FRVI (30k) & 0.477 & $1.67\times10^{-3}$ & 0.476 & 47.9 \\
\midrule
Heterogeneous, N=8 & AFIN & 0.320 & $5.89\times10^{-4}$ & 0.490 & 11.1 \\
 & FRVI (1k) & 0.402 & $1.13\times10^{-3}$ & 0.525 & 38.9 \\
 & FRVI (10k) & 0.389 & $8.30\times10^{-4}$ & 0.531 & 41.7 \\
 & FRVI (30k) & 0.398 & $1.43\times10^{-3}$ & 0.529 & 41.7 \\
\midrule
Heterogeneous, N=32 & AFIN & 0.247 & $1.74\times10^{-4}$ & 0.484 & 0.0 \\
 & FRVI (1k) & 0.409 & $1.82\times10^{-4}$ & 0.432 & 0.0 \\
 & FRVI (10k) & 0.477 & $1.42\times10^{-4}$ & 0.454 & 0.0 \\
 & FRVI (30k) & 0.481 & $2.11\times10^{-4}$ & 0.450 & 0.0 \\
\bottomrule
\end{tabular}
\end{table}

\subsection{Calibration}
\label{app:calibration}

We assess posterior uncertainty through marginal-interval coverage (\Cref{app:metrics}).
The study uses 200 simulated tasks spanning all prior configurations and likelihood forms, including mixed models: 120 inside the training ranges, 40 with \(N\) from 384 to \(1{,}024\), and 40 with \(D\) from 24 to 64.
Inside the training ranges, nominal 50, 80, 90, and 95\% intervals cover \(0.47\), \(0.77\), \(0.875\), and \(0.93\) of the true coordinates, a mean absolute deviation of \(2.6\) percentage points.
SNIS with \(4{,}096\) proposals gives similar coverage (\(0.871\) at the 90\% level).

On a separate 64-task panel with \(N\) between 320 and 400, 90\% coverage is \(0.907\).
In the larger out-of-range panels it is \(0.57\) for \(N\) from 384 to \(1{,}024\) and \(0.74\) for \(D\) from 24 to 64.
On the conjugate tasks of \Cref{app:references}, whose posteriors are known exactly, 90\% coverage is \(0.92\) for direct predictions and \(0.89\) after SNIS.

\subsection{Forward time and memory}
\label{app:scaling}

\Cref{tab:scaling} measures one warmed forward pass of the trained model on a single GPU, as the median over synchronized repetitions, together with peak GPU memory.
Time is flat at about \(10\) ms up to \(D=16\) or \(N=64\) and then grows with the pair arrays and the attention scores, as \Cref{app:complexity} predicts, reaching \(67\) ms and \(1.1\) GB at \(D=64\) and \(71\) ms and \(1.3\) GB at \(N=1{,}024\).
The first call on a fresh process is slower, about \(0.5\) s, because of kernel compilation.

\begin{table}[!htbp]
\centering
\caption{Warmed forward-pass time and peak GPU memory of the trained model. Left: \(D\) varies at \(N=64\); right: \(N\) varies at \(D=16\).}
\label{tab:scaling}
\small
\begin{tabular}{@{}rrr@{\qquad}rrr@{}}
\toprule
\(D\) & Time (ms) & Memory (MB) & \(N\) & Time (ms) & Memory (MB) \\
\midrule
4  & 9.8  & 54    & 1     & 9.8  & 38    \\
8  & 9.8  & 64    & 16    & 9.9  & 49    \\
16 & 9.5  & 101   & 64    & 9.1  & 101   \\
24 & 13.2 & 187   & 256   & 17.8 & 316   \\
32 & 17.7 & 308   & 512   & 35.3 & 618   \\
48 & 38.9 & 650   & 1,024 & 71.4 & 1,272 \\
64 & 66.7 & 1,130 &       &      &       \\
\bottomrule
\end{tabular}
\end{table}

\subsection{Poisson extension}
\label{app:poisson}

\paragraph{Tasks.}
The new likelihood is \(y_n\mid z\sim\mathrm{Poisson}(\exp(x_n^\top z))\), with the four pretraining prior configurations and covariates scaled by \(0.5\) to keep the rates moderate; all likelihood factors in these tasks are Poisson.
The held-out set has 24 tasks, two for each \(D\in\{2,4,8,16\}\) and \(N\in\{16,64,256\}\).
References are four-chain NumPyro runs as in \Cref{app:references}; M1 and M2 use all reference draws, and SW\(_2\) uses \(4{,}096\) of them against \(4{,}096\) samples or proposals.
These tasks are a separate set, so their absolute errors are not comparable with the benchmark curves.

\paragraph{Adapter.}
The Poisson adapter follows \Cref{app:encoder}.
Its node descriptor has five entries, \(y_n\), \(\log(1+y_n)\), \(\log(1+\|x_n\|_2)\), \(x_{ni}\), and \(y_nx_{ni}\); its pair descriptor has nine, the same three scalars, \(x_{ni}\), \(x_{nj}\), their product, both count--covariate products, and a diagonal indicator.
With one hidden layer of width 64 and \(C=40\) output features it has \(6{,}224\) parameters.

\paragraph{Training.}
Both modes start from the pretrained AFIN and run \(30{,}000\) AdamW steps (batch size 32, four accumulated micro-batches, weight decay \(10^{-4}\), 500 warmup steps, cosine decay) on tasks with \(D\) uniform in \(\{1,\ldots,16\}\) and \(N\) in \(\{1,\ldots,256\}\), cycling through the prior configurations and covariate generators.
Adapter-only training uses learning rate \(10^{-3}\) and an exponential moving average of the adapter weights; full fine-tuning uses \(2\times10^{-4}\) and the final weights.
Each mode ran three seeds, all reported.
After adapter-only training the shared network is unchanged, so every original factor type behaves exactly as before.
\Cref{tab:poisson-summary,tab:poisson-seeds,tab:poisson-additional} give SW\(_2\), the per-seed results, and mean error, covariance error, and 90\% coverage.

\begin{table}[!htbp]
\centering
\caption{Poisson extension: SW$_2$ and training time.
    Entries are means $\pm$ sample standard deviations over three training seeds; each accuracy value first averages over 24 held-out tasks.
    SNIS uses 4,096 proposals. Best in bold.}
\label{tab:poisson-summary}
\small
\begin{tabular}{@{}lrrr@{}}
\toprule
& \multicolumn{2}{c}{Posterior error (SW$_2$)} & Training time \\
\cmidrule(lr){2-3}
Training mode & Direct & +SNIS & (GPU-h) \\
\midrule
Adapter only & $0.05910\pm0.00198$ & $\mathbf{0.03011}\pm0.00019$ & $\mathbf{4.946}\pm0.015$ \\
Full fine-tuning & $\mathbf{0.04580}\pm0.00055$ & $0.03048\pm0.00036$ & $8.392\pm0.073$ \\
\bottomrule
\end{tabular}
\end{table}

\begin{table}[!htbp]
\centering
\caption{Poisson extension by training seed: mean SW\(_2\) over the same 24 tasks.}
\label{tab:poisson-seeds}
\small
\begin{tabular}{lrrrr}
\toprule
& \multicolumn{2}{c}{Adapter only} & \multicolumn{2}{c}{Full fine-tuning} \\
\cmidrule(lr){2-3}\cmidrule(lr){4-5}
Seed & Direct & +SNIS & Direct & +SNIS \\
\midrule
Seed 1 & 0.06139 & 0.03029 & 0.04606 & 0.03071 \\
Seed 2 & 0.05801 & 0.03013 & 0.04518 & 0.03067 \\
Seed 3 & 0.05789 & 0.02991 & 0.04618 & 0.03007 \\
\bottomrule
\end{tabular}
\end{table}

\begin{table}[!htbp]
\centering
\caption{Additional Poisson results, averaged over tasks within each training
    seed and then reported as mean $\pm$ sample standard deviation across three
    seeds. Mean and covariance errors use the Euclidean and Frobenius norms.
    Coverage is the fraction of simulated latent coordinates within nominal
    90\% marginal intervals. Best in bold; for coverage, closest to the nominal 90\%.}
\label{tab:poisson-additional}
\small
\begin{tabular}{llccc}
\toprule
Training & Inference & Mean error & Covariance error & 90\% coverage \\
\midrule
Adapter only & Direct & $0.1166\pm0.0039$ & $0.1565\pm0.0034$ & $\mathbf{0.8993}\pm0.0030$ \\
Adapter only & +SNIS & $\mathbf{0.0319}\pm0.0005$ & $\mathbf{0.1135}\pm0.0035$ & $0.9306\pm0.0015$ \\
Full fine-tuning & Direct & $0.0785\pm0.0014$ & $0.1401\pm0.0065$ & $0.9219\pm0.0000$ \\
Full fine-tuning & +SNIS & $0.0331\pm0.0006$ & $0.1147\pm0.0031$ & $0.9314\pm0.0015$ \\
\bottomrule
\end{tabular}
\end{table}

\subsection{Ablations}
\label{app:ablation}

The variants of \Cref{tab:ablation} keep AFIN's adapters and Gaussian decoder.
The node-only variants drop the pair array everywhere, so the decoder predicts the precision from node features alone.
The sum variants remove the Transformer blocks and pass the sum of the encoded factors straight to the decoder, which is the conjugate template of \Cref{sec:architecture} with a learned encoder and decoder.
\Cref{tab:ablation-full} gives the direct results at full precision, with one standard deviation of SW\(_2\) across evaluation repetitions.

\begin{table}[!htbp]
\centering
\caption{Ablations on the synthetic suite, direct prediction; best in bold.}
\label{tab:ablation-full}
\small
\begin{tabular}{@{}lrrr@{}}
\toprule
Variant & M1 & M2 & SW$_2$ \\
\midrule
Node and pair, attention (AFIN) & \textbf{0.0533} & \textbf{0.1047} & $\mathbf{0.0412}\pm0.0003$ \\
Node only, attention & 0.0734 & 0.2078 & $0.0490\pm0.0003$ \\
Node and pair, sum & 0.2903 & 0.2484 & $0.1324\pm0.0009$ \\
Node only, sum & 0.3140 & 0.2758 & $0.1425\pm0.0008$ \\
\bottomrule
\end{tabular}
\end{table}

\subsection{Additional ablations and factor composition}
\label{app:additional-ablations}

These experiments test whether AFIN can combine factors in ways not seen during training, and what learned attention contributes beyond fixed mixing.
All results are direct predictions, without SNIS.
Tables report means and sample standard deviations over three training seeds, after averaging tasks within each seed.

The prior--likelihood and merge comparisons use three newly trained networks per variant, with 100,000 steps and final exponential-moving-average weights, following the optimization settings of \Cref{sec:experimental-setup}.
These networks and their SW\(_2\) evaluation protocol differ from those of \Cref{tab:ablation-full}; scoring details are in \Cref{app:metrics}.

\paragraph{Unseen prior--likelihood combinations.}
We compare networks trained on all 16 prior--likelihood combinations with networks trained on 12.
The excluded pairings are diagonal Gaussian prior with Gaussian likelihood, full-covariance Gaussian prior with Bernoulli-logit likelihood, independent Student-\(t\) prior with Binomial-logit likelihood, and independent Laplace prior with Student-\(t\) likelihood.
Both groups sample individual prior configurations and likelihood types uniformly, and each training task uses one likelihood type throughout.
On the 144 homogeneous synthetic tasks, the retained combinations account for 108 tasks and the excluded combinations for 36.
\Cref{tab:composition-control} shows little change on retained combinations, with higher errors on excluded combinations.

\begin{table}[H]
\centering
\small
\setlength{\tabcolsep}{4pt}
\caption{Generalization to excluded prior--likelihood combinations. ``Retained'' and ``excluded'' refer to the model trained on 12 combinations; the model trained on all 16 sees both groups.}
\label{tab:composition-control}
\begin{tabular}{@{}llrrr@{}}
\toprule
Evaluation tasks & Training & M1 & M2 & SW$_2$ \\
\midrule
Retained (108) & All 16 & $0.0488\pm0.0011$ & $0.1171\pm0.0111$ & $0.0278\pm0.0002$ \\
 & 12 only & $0.0499\pm0.0012$ & $0.1166\pm0.0120$ & $0.0279\pm0.0003$ \\
\addlinespace[3pt]
Excluded (36) & All 16 & $0.0521\pm0.0046$ & $0.1135\pm0.0089$ & $0.0284\pm0.0013$ \\
 & 12 only & $0.0716\pm0.0047$ & $0.1364\pm0.0098$ & $0.0347\pm0.0018$ \\
\bottomrule
\end{tabular}
\end{table}

\paragraph{Combining a new likelihood with existing types.}
The three adapter-only checkpoints of \Cref{app:poisson} were trained on Poisson-only tasks.
We evaluate them without further training on Poisson-only controls and mixtures of Poisson with Gaussian, Bernoulli-logit, or Student-\(t\) likelihoods.
In each mixture, half the observations use each type.
Crossing four prior configurations, \(D\in\{4,8\}\), \(N\in\{32,128\}\), and two task seeds gives 32 tasks per group, or 128 in total.
Each control and its mixed variants share the prior, latent value, and covariates, with the same covariate scaling as adapter training.

\Cref{tab:poisson-composition} reports standardized mean error, relative covariance error, and posterior mass inside the predicted 90\% ellipsoid.
Averaged over tasks, this mass remains close to 90\% in all four groups, supporting use of the new adapter alongside existing likelihood types.
Reference construction and metric definitions are given in \Cref{app:references,app:metrics}.

\begin{table}[H]
\centering
\small
\setlength{\tabcolsep}{4pt}
\caption{Poisson composition without training on mixtures. Each entry averages 32 tasks within each of three adapter-training seeds.}
\label{tab:poisson-composition}
\begin{tabular}{@{}lrrr@{}}
\toprule
Likelihoods & \makecell{Standardized\\mean error} & \makecell{Relative\\covariance error} & \makecell{Posterior mass\\in 90\% region} \\
\midrule
Poisson only & $0.0728\pm0.0009$ & $0.0771\pm0.0014$ & $0.8947\pm0.0008$ \\
Poisson + Gaussian & $0.0973\pm0.0037$ & $0.0860\pm0.0038$ & $0.8952\pm0.0032$ \\
Poisson + Bernoulli-logit & $0.0767\pm0.0076$ & $0.1152\pm0.0095$ & $0.9133\pm0.0030$ \\
Poisson + Student-$t$ & $0.0969\pm0.0061$ & $0.1098\pm0.0024$ & $0.9019\pm0.0027$ \\
\bottomrule
\end{tabular}
\end{table}

\paragraph{Learned attention versus fixed mixing.}
We keep the encoder, decoder, and merge-block structure, and change how each block combines factor values.
\emph{Uniform mixing} averages the value features over all factors; \emph{self-only mixing} uses only each factor's own value features until the final sum.
Both retain the value and output projections, residual connections, and feed-forward blocks.
Their query/key projections and attention-score weights are inactive, giving them fewer trainable parameters than learned attention.
Uniform mixing improves all three average errors over self-only mixing (\Cref{tab:controlled-merge}), showing the benefit of interaction before the final sum.
Learned attention further improves M1 and SW\(_2\), while average M2 is essentially unchanged.

\begin{table}[H]
\centering
\small
\setlength{\tabcolsep}{4pt}
\caption{Merge comparison on the same 144 homogeneous synthetic tasks. All three variants retain node and pair features. Standard deviations are across independently trained networks.}
\label{tab:controlled-merge}
\begin{tabular}{@{}lrrr@{}}
\toprule
Merge & M1 & M2 & SW$_2$ \\
\midrule
Learned attention & $0.0558\pm0.0015$ & $0.1238\pm0.0106$ & $0.0298\pm0.0001$ \\
Uniform mixing & $0.0732\pm0.0055$ & $0.1238\pm0.0043$ & $0.0350\pm0.0020$ \\
Self-only mixing & $0.2903\pm0.0008$ & $0.2310\pm0.0081$ & $0.1211\pm0.0002$ \\
\bottomrule
\end{tabular}
\end{table}

\subsection{Amortized baseline}
\label{app:npe-pfn}

NPE-PFN \citep{vetter2025npepfn} feeds prior-predictive simulations of a task to a tabular foundation model as context and reads the posterior off its in-context predictions, so it needs no task-specific training.
We ran the official implementation with frozen TabPFN v2 weights \citep{hollmann2025tabpfnv2} on the 144 synthetic tasks, with 1,000 or 10,000 prior-predictive simulations as context and 2,048 posterior samples per task, scored against the references of \Cref{app:references}; the 1,000-simulation context is a prefix of the 10,000.
AFIN's moments are closed-form and NPE-PFN's are estimated from its samples; SW\(_2\) uses 2,048 samples for both.
This separate sample-based comparison retains its own evaluation budget and timing, rather than the Gaussian-quantile scoring and warmed forward-pass timing of \Cref{fig:synth-posterior-time}.
Times are per task for inference and sampling, excluding the context simulations.
\Cref{tab:npe-pfn-full} splits the results by setting.
AFIN has the lowest error in every block; the larger context reduces NPE-PFN's errors but costs 115 s per task.
With the same 2,048-proposal SNIS correction applied to both, AFIN keeps the lower mean and SW\(_2\) errors and a higher effective sample size, 0.81 against 0.60 of the proposals.

\begin{table}[!htbp]
\centering
\caption{NPE-PFN against AFIN on the homogeneous synthetic suite by setting; best in bold.}
\label{tab:npe-pfn-full}
\small
\begin{tabular}{@{}llrrr@{}}
\toprule
Tasks & Method & M1 & M2 & SW$_2$ \\
\midrule
All (144) & AFIN & \textbf{0.0533} & \textbf{0.1047} & \textbf{0.0415} \\
 & NPE-PFN, 1,000 context simulations & 0.2417 & 0.3008 & 0.1133 \\
 & NPE-PFN, 10,000 context simulations & 0.1298 & 0.2542 & 0.0761 \\
\midrule
Data-rich (48) & AFIN & \textbf{0.0238} & \textbf{0.0039} & \textbf{0.0158} \\
 & NPE-PFN, 1,000 & 0.1691 & 0.0720 & 0.1038 \\
 & NPE-PFN, 10,000 & 0.0923 & 0.0421 & 0.0628 \\
\midrule
Intermediate (48) & AFIN & \textbf{0.0644} & \textbf{0.0395} & \textbf{0.0353} \\
 & NPE-PFN, 1,000 & 0.3453 & 0.1111 & 0.1281 \\
 & NPE-PFN, 10,000 & 0.1848 & 0.0875 & 0.0723 \\
\midrule
Data-scarce (48) & AFIN & \textbf{0.0716} & \textbf{0.2707} & \textbf{0.0734} \\
 & NPE-PFN, 1,000 & 0.2106 & 0.7193 & 0.1079 \\
 & NPE-PFN, 10,000 & 0.1124 & 0.6331 & 0.0932 \\
\bottomrule
\end{tabular}
\end{table}

\clearpage
\section{Extended Related Work}
\label{app:related}

\paragraph{Expressive posterior approximators.}
Normalizing flows give tractable, invertible maps from a simple base density to flexible posteriors \citep{pmlr-v37-rezende15,papamakarios2021normalizing}.
Coupling and autoregressive designs such as RealNVP, IAF, MAF, and neural spline flows \citep{dinh2017density,kingma2016iaf,papamakarios2017maf,durkan2019nsf,agrawal2024disentangling,ko2025modelinformedflows} serve both as variational families and as conditional density estimators in simulation-based inference \citep{NIPS2016_6aca9700,pmlr-v97-greenberg19a,pmlr-v89-papamakarios19a,radev2022bayesflow}.
Diffusion and score-based models have been adapted to posterior sampling for inverse problems \citep{chung2023dps} and to simulation-based inference by learning or composing posterior scores \citep{pmlr-v202-geffner23a,pmlr-v235-sharrock24a}, and flow matching trains continuous flows by vector-field regression \citep{lipman2023flowmatching,tong2024cfm,albergo2023stochasticinterpolants}, including for scalable simulation-based inference \citep{dax2023fmpe}.
All of these operate within a fixed model, simulator, or task interface; they are complementary to AFIN's factor interface, which could feed a more expressive decoder.

\paragraph{Amortization over model families and programs.}
BayesFlow learns globally amortized posterior estimators with invertible networks and learned summaries for a fixed simulator family \citep{radev2022bayesflow}, and JANA jointly amortizes the posterior and the likelihood, which also gives marginal likelihoods and posterior predictives \citep{radev2023jana}.
SA-ABI shares weights across alternative prior, likelihood, and data choices for sensitivity analysis \citep{elsemueller2024sensitivity}.
Prior-fitted networks learn Bayesian prediction over supervised tasks drawn from a task prior and target predictive rather than latent posteriors \citep{muller2022pfn}; ACE conditions on data, latents, and priors through a general token interface \citep{chang2025ace}; inference compilation trains proposals for a probabilistic program and uses them inside sequential importance sampling \citep{le2017inference}; and meta-amortized VI shares inference computation across related generative models \citep{wu2020metaamortized}.
None of these exposes the inference problem as a typed factor list with dimension-independent decoding, which is what lets one AFIN cover every column of \Cref{tab:main-related} without per-task fitting.

\paragraph{Message passing.}
The observation behind our merge, that the useful summary of one factor depends on where the other factors put their mass, is also the basis of expectation propagation \citep{minka2001ep}, which refines each factor's approximation against the product of the others.
AFIN replaces this iterative, model-specific computation with one learned exchange of information across factors, amortized over models.

\clearpage

\end{document}